\documentclass[]{template}

\setboolean{logo}{true}

\ifXeTeX
  \usepackage{fontspec}
  \AtEndPreamble{%
    \setmainfont{XCharter-Roman.otf}[
      BoldFont=XCharter-Bold.otf,
      ItalicFont=XCharter-Italic.otf,
      BoldItalicFont=XCharter-BoldItalic.otf,
      Ligatures=TeX
    ]%
    \microtypesetup{tracking=false}%
    \normalfont
  }
\fi

\usepackage{amsmath,amsfonts,bm}

\def\eqref#1{equation~\ref{#1}}

\def\1{\bm{1}}

\DeclareMathAlphabet{\mathsfit}{\encodingdefault}{\sfdefault}{m}{sl}
\SetMathAlphabet{\mathsfit}{bold}{\encodingdefault}{\sfdefault}{bx}{n}

\usepackage{float}
\usepackage[section]{placeins}
\usepackage{array}
\usepackage{wrapfig}
\usepackage{titletoc}
\usepackage{tikz}
\usetikzlibrary{arrows.meta,positioning,fit,calc,shapes.geometric}

\definecolor{tbInk}{HTML}{2C3E48}
\definecolor{tbBlue}{HTML}{5D5D5D}
\definecolor{tbBlueLight}{HTML}{E8F1F6}
\definecolor{tbTeal}{HTML}{3D7B75}
\definecolor{tbTealLight}{HTML}{E8F3F1}
\definecolor{tbAmber}{HTML}{B77A2C}
\definecolor{tbAmberLight}{HTML}{FAF0E1}
\definecolor{tbGrayLight}{HTML}{F2F4F5}

\definecolor{schedShuffle}{HTML}{6B7280}
\definecolor{schedForwardGap}{HTML}{147D73}
\definecolor{schedReverseGap}{HTML}{4C63B6}
\definecolor{schedForwardMix}{HTML}{B56500}
\newcommand{\schedshuffle}{\textcolor{schedShuffle}{\textsc{Shuffle}}}
\newcommand{\schedforwardgap}{\textcolor{schedForwardGap}{\textsc{Forward--Gap}}}
\newcommand{\schedreversegap}{\textcolor{schedReverseGap}{\textsc{Reverse--Gap}}}
\newcommand{\schedforwardmix}{\textcolor{schedForwardMix}{\textsc{Forward--Mix}}}

\title{Scaling Mathematical Proof Supervision from Structured Textbooks}

\author{%
\normalfont\bfseries%
Youqing Fang$^{1,3*}$,
Zhouqi Hua$^{2,3*}$,
Yinhao Tang$^{1,3}$,
Yuzhe Gu$^{3,4}$,
Lingkai Kong$^{3,4}$, \authorcr
\bfseries%
Bin Liu$^{1}$,
Weiming Zhang$^{1}$,
Kai Chen$^{3}$,
Yanan Sun$^{3 \dagger}$,
Wenwei Zhang$^{3 \dagger}$ \\
\normalfont%
$^{1}$University of Science and Technology of China \quad
$^{2}$Fudan University \\
$^{3}$Shanghai AI Laboratory \quad 
$^{4}$Shanghai Jiao Tong University
}

\usepackage{etoolbox}
\patchcmd{\maketitle}
  {\begin{adjustwidth}{0pt}{24pt}}
  {\begin{adjustwidth}{0pt}{0pt}}
  {}
  {\PackageError{intern-layout}{Unable to patch title width}{Check template.cls}}
\hypersetup{
  colorlinks=true,
  linkcolor=red,
  citecolor=Cerulean,
  filecolor=magenta,
  urlcolor=magenta
}
\renewcommand{\paragraph}[1]{\vspace{1mm}\noindent\textbf{#1}}

\begin{abstract}

Mathematical supervision for language models is typically flattened into isolated problem--solution pairs, discarding the curricular organization and evidential structure of the original sources.
We introduce \textbf{TOME-Math} (\textbf{T}extbook-\textbf{O}riented \textbf{M}athematical \textbf{E}vidence), a large-scale corpus that converts 2,118 textbooks spanning the core areas of university-level mathematics subjects into approximately 380K structured proof-supervision records while preserving chapter order, proposition--proof links, dependency references, and source provenance.
Leveraging these structural signals, we explore curriculum learning across multiple mathematical subjects and characterize scaling patterns with supervision volume and source breadth.
At full scale, fine-tuning on TOME-Math raises the unweighted mean score across our held-out evaluations and established public benchmarks by \textbf{16.7} percentage points over the base model, while exceeding all three public instruction-data baselines evaluated in aggregate.
Our findings highlight structured textbooks as both a scalable source of mathematical supervision and an empirical framework for studying the impact of data organization on mathematical reasoning.
We will release the corpus to facilitate further research on structured mathematical supervision.

\par\smallskip
\noindent\textbf{Project page:}\quad
\textcolor[HTML]{2176C1}{https://tome-math.github.io/}
\end{abstract}

\begin{document}

\maketitle
\begingroup
\renewcommand{\thefootnote}{\fnsymbol{footnote}}
\footnotetext[1]{Equal contribution, $^{\dagger}$Corresponding authors.}
\endgroup

\begin{figure}[h]
    \centering
    \includegraphics[width=\linewidth]{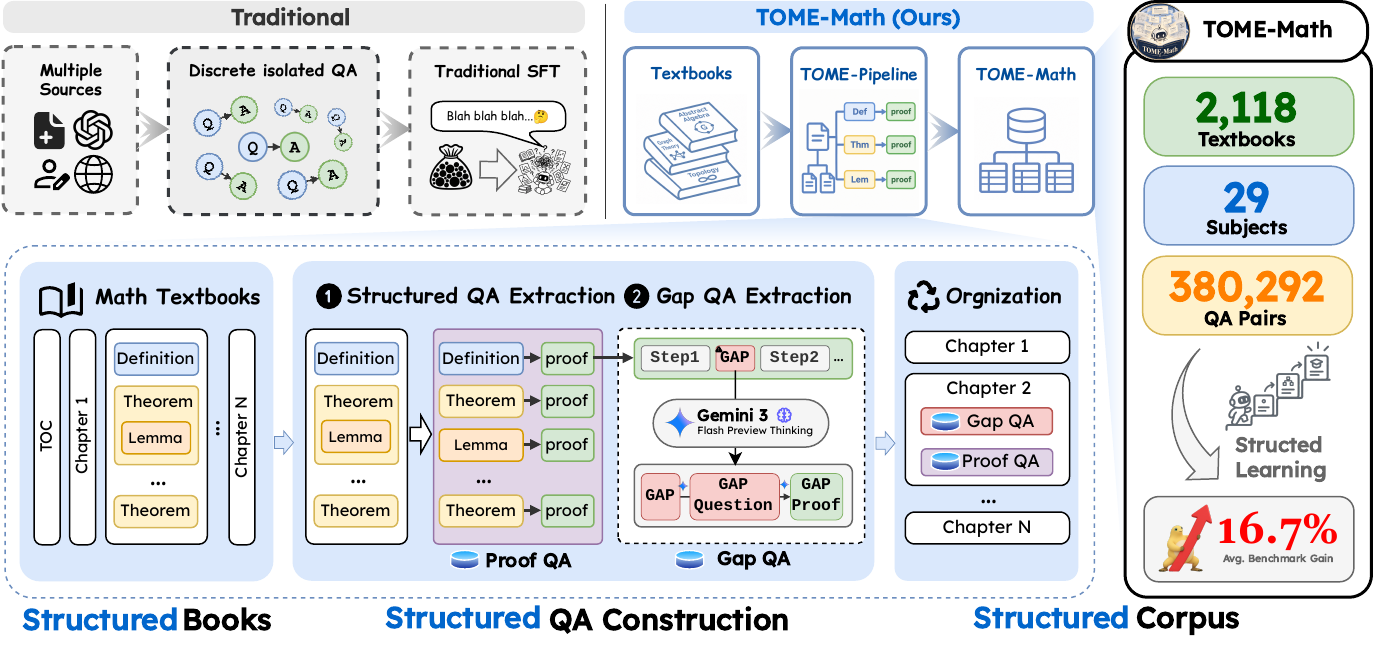}
    \caption{Overview of TOME-Math. \textbf{Top-left}: Traditional pipelines reduce diverse sources to isolated QAs. \textbf{Bottom-left}: TOME-Math preserves textbook structure through structured QA construction and curriculum-based SFT. \textbf{Right}: TOME-Math comprises 380,292 QAs from 2,118 textbooks spanning 29 subjects.}
    \label{fig:teaser}
\end{figure}

\section{Introduction}

Mathematical reasoning in large language models has advanced rapidly, driven in part by improvements in training data and in part by advances in reasoning-time computation \citep{shao2024deepseekmath,yang2024qwen25math,openai2024o1,guo2025deepseekr1,yang2025qwen3,bai2026intern,shao2025deepseekmath,zhao2026achieving,gao2025long,tang2026next, hua2026imitation, anonymous2026incentivizing,anonymous2026inverseharness}. On the data side, recent efforts have expanded mathematical supervision to millions of QA pairs (\textit{i.e.}, Question-Answer pairs) and in doing so have revealed that properties such as question diversity, solution format, and teacher-model quality significantly shape downstream performance \citep{yu2023metamath,yue2024mammoth,toshniwal2024openmathinstruct,kong2026advancedmathbench}. These findings suggest a broader principle: \textit{The value of mathematical supervision is not determined by the number of training examples alone, but also by how those examples are constructed and organized}.

Despite this, the dominant paradigm for mathematical supervision remains the isolated problem--solution pair. Whether the original knowledge comes from competition archives, grade-school word problems, web exercises, or synthetic augmentation, the resulting training data typically takes the uniform shape of $\text{problem} \to \text{solution}$ \citep{cobbe2021training,hendrycks2021math,yu2023metamath}. This representation is convenient to scale, but it does not explicitly leverage the organizational structure that many source materials already possess.


Mathematical textbooks are a notable case in point.
A textbook is not merely a collection of solved problems, but a carefully structured body of knowledge. Definitions introduce core concepts, lemmas and theorems develop them, proofs establish their validity, and exercises reinforce understanding. These elements are organized within a chapter hierarchy that reflects prerequisite relations, conceptual dependencies, and pedagogical progression.
Prior work has explored textbooks as a source of mathematical pretraining data, problem--solution pairs, and proof supervision \citep{wang2023mathpile,tang2024mathscale,welleck2021naturalproofs,azerbayev2023proofnet,huang2025formarl,lee2026leangap,zhang2026scalable}. However, textbooks are often treated primarily as sources of mathematical content, leaving their native structure underexplored. In particular, it remains unclear how to scale natural-language proof supervision while preserving chapter order, proposition dependencies, source locations, and exercises in a unified framework.


To address this gap, we introduce \textbf{TOME-Math} (\textbf{T}extbook-\textbf{O}riented \textbf{M}athematical \textbf{E}vidence), a large-scale proof-supervision resource constructed from 2{,}118 mathematics textbooks. Central to TOME-Math is the notion of \textit{structured} mathematical supervision, which preserves structure at three levels. First, \textbf{hierarchical structure} retains the native organization of textbooks, including chapters, propositions, and their ordering. Second, \textbf{dependency structure} connects theorem-level proof questions with the intermediate proof obligations derived from gaps in their proofs. Third, \textbf{curriculum structure} organizes supervision according to textbook order and proof granularity, enabling controlled training over related examples rather than independent QA pairs. Following these principles, TOME-Math contains 380{,}292 proof-oriented QA pairs spanning 29 mathematical subjects. Figure~\ref{fig:teaser} provides an overview of the construction and organization of the resulting corpus.


We further validate the effectiveness of TOME-Math through supervised fine-tuning (SFT) on a 30B base model~\citep{yang2025qwen3}. Its retained structure enables controlled studies of how the organization of supervision affects mathematical reasoning. Rather than treating scale as the only variable, TOME-Math allows us to examine the roles of retrieval structure, curriculum order, and source coverage under controlled settings. Empirically, TOME-Math SFT improves the six-benchmark average from 29.17\% to 45.83\%, while further experiments show consistent benefits from structure-guided retrieval, textbook-aligned curricula, and broader source coverage. Our contributions are:

\begin{itemize}
    \item We introduce TOME-Math, a large-scale proof-supervision resource
    constructed from 2{,}118 mathematics textbooks. Through a
    structure-preserving pipeline, it produces 380{,}292 Concept and Gap QA
    records while retaining chapter progression, proposition--proof links,
    dependency references, source provenance, and a separate inventory of
    textbook-authored exercises.
    \item We demonstrate the overall effectiveness of TOME-Math through
    full-corpus fine-tuning. On book-held-out textbook exercises and established
    public benchmarks, TOME-Math achieves the best performance on five of six
    evaluations, improving the unweighted six-benchmark mean from 29.17\% to
    45.83\% over the base model and outperforming all three public
    instruction-data baselines in aggregate.
    \item We systematically investigate how textbook structure contributes to
    mathematical reasoning at inference and training time. Structure-guided
    retrieval improves proof-generation pass@1 from 13.25\% to 17.38\% under
    a matched context budget; across six subjects and three data scales,
    forward textbook curricula consistently outperform shuffled and reverse
    schedules; and broader source coverage yields further gains under a fixed
    training-token budget.
\end{itemize}

\begin{table}[t]
    \centering
    \scriptsize
    \setlength{\tabcolsep}{2.5pt}
    \renewcommand{\arraystretch}{1.18}
    \caption{Mathematical resources by scale, supervision, and structure.
    Sizes retain reported units; sources include synthetic seeds.
    NL = natural language; refs = proposition references.
    }
    \label{tab:related-datasets}
    \resizebox{\linewidth}{!}{%
    \begin{tabular}{@{}lrlll@{}}
        \hline
        \textbf{Resource} & \textbf{Size} & \textbf{Source} &
        \textbf{Supervision} & \textbf{Metadata / structure} \\
        \hline
        \rowcolor{gray!15}[0pt][0pt]
        \multicolumn{5}{@{}l@{}}{\textit{Mathematical instruction data}} \\
        MATH~\citeyearpar{hendrycks2021math} & 12.5K QA & Competitions & Solutions & Subject, difficulty \\
        MetaMathQA~\citeyearpar{yu2023metamath} & 395K QA & GSM8K, MATH & Solutions & Seeds, augmentation types \\
        OpenMathInstruct-2~\citeyearpar{toshniwal2024openmathinstruct} & 14M QA & GSM8K, MATH & Solutions & Source labels \\
        NuminaMath-CoT~\citeyearpar{li2024numinamath} & ${\sim}$860K QA & Exams, web, synthetic & Solutions & Source labels \\
        \rowcolor{gray!15}[0pt][0pt]
        \multicolumn{5}{@{}l@{}}{\textit{Textbook and knowledge-informed resources}} \\
        MathPile~\citeyearpar{wang2023mathpile} & 9.5B tokens & Books, papers, web & Text & Sources, file paths \\
        CollegeMath~\citeyearpar{tang2024mathscale} & 4,099 QA & 9 textbooks & Short answers & Book/section/exercise IDs \\
        \rowcolor{gray!15}[0pt][0pt]
        \multicolumn{5}{@{}l@{}}{\textit{Proof resources and evaluation}} \\
        NaturalProofs~\citeyearpar{welleck2021naturalproofs} & 32K proofs & 2 textbooks + web & NL proofs & Reference graph \\
        ProofNet~\citeyearpar{azerbayev2023proofnet} & 371 examples & Textbooks, Putnam & NL + Lean stmt. & Book/exercise IDs \\
        Lean-GAP~\citeyearpar{lee2026leangap} & 430 problems & 1 textbook & Lean statements & Section/exercise IDs \\
        AlgGeoTest~\citeyearpar{peng2025proof2hybrid} & 456 questions & Stacks Project & Proof judgments & Statement tags \\
        \hline
        \textbf{TOME-Math (ours)} & \textbf{380,292 QA} & \textbf{2,118 textbooks} &
        \textbf{NL + local proofs} & \textbf{Chapter order, refs, provenance} \\
        \hline
    \end{tabular}}
\end{table}

\section{Related Work}
\label{sec:related-work}


\paragraph{Mathematical instruction data.}
GSM8K and MATH provide human-authored problems and solutions for grade-school and competition mathematics \citep{cobbe2021training,hendrycks2021math}. MetaMathQA and OpenMathInstruct-2 scale these seeds through synthetic augmentation \citep{yu2023metamath,toshniwal2024openmathinstruct}, while NuminaMath-CoT aggregates diverse sources with chain-of-thought solutions \citep{li2024numinamath}. Their primary supervision unit is a problem--solution pair, accompanied by source, seed-question, or difficulty metadata. Table~\ref{tab:related-datasets} contrasts this organization with the textbook hierarchy and proposition links retained by TOME-Math.


\paragraph{Textbook and knowledge-informed resources.}
OpenWebMath and MathPile provide mathematical pretraining text, with MathPile incorporating textbooks alongside papers and web sources \citep{paster2023openwebmath,wang2023mathpile}. CollegeMath extracts textbook exercises and short answers, while MathScale synthesizes QA using a graph of topics and knowledge points \citep{tang2024mathscale}. TheoremQA evaluates theorem application across STEM subjects \citep{chen2023theoremqa}. These efforts motivate retaining both textbook organization and links to mathematical statements when constructing proof supervision.

\paragraph{Proof resources and evaluation.}
NaturalProofs preserves statements, proofs, and a reference graph \citep{welleck2021naturalproofs,welleck2022naturalprover}, while NaturalProofs-Gen and NaturalProver study reference-grounded proof generation . ProofNet and Lean-GAP connect textbook mathematics to formalization \citep{azerbayev2023proofnet,lee2026leangap}, while FormaRL's uproof provides 5,273 proof problems from 14 textbooks for autoformalization evaluation \citep{huang2025formarl}. Proof2Hybrid's AlgGeoTest studies proof-derived judgments, and the Open Proof Corpus studies automated assessment of human-evaluated model proofs \citep{peng2025proof2hybrid,dekoninck2025openproofcorpus}. Building on these directions, TOME-Math combines coverage of 2,118 textbooks with self-contained local proof obligations, chapter order, and proposition references, supporting controlled studies of source coverage, proof-target format, and training order.

\section{TOME-Math}

\begin{figure}[t]
\centering

\includegraphics[width=\linewidth]{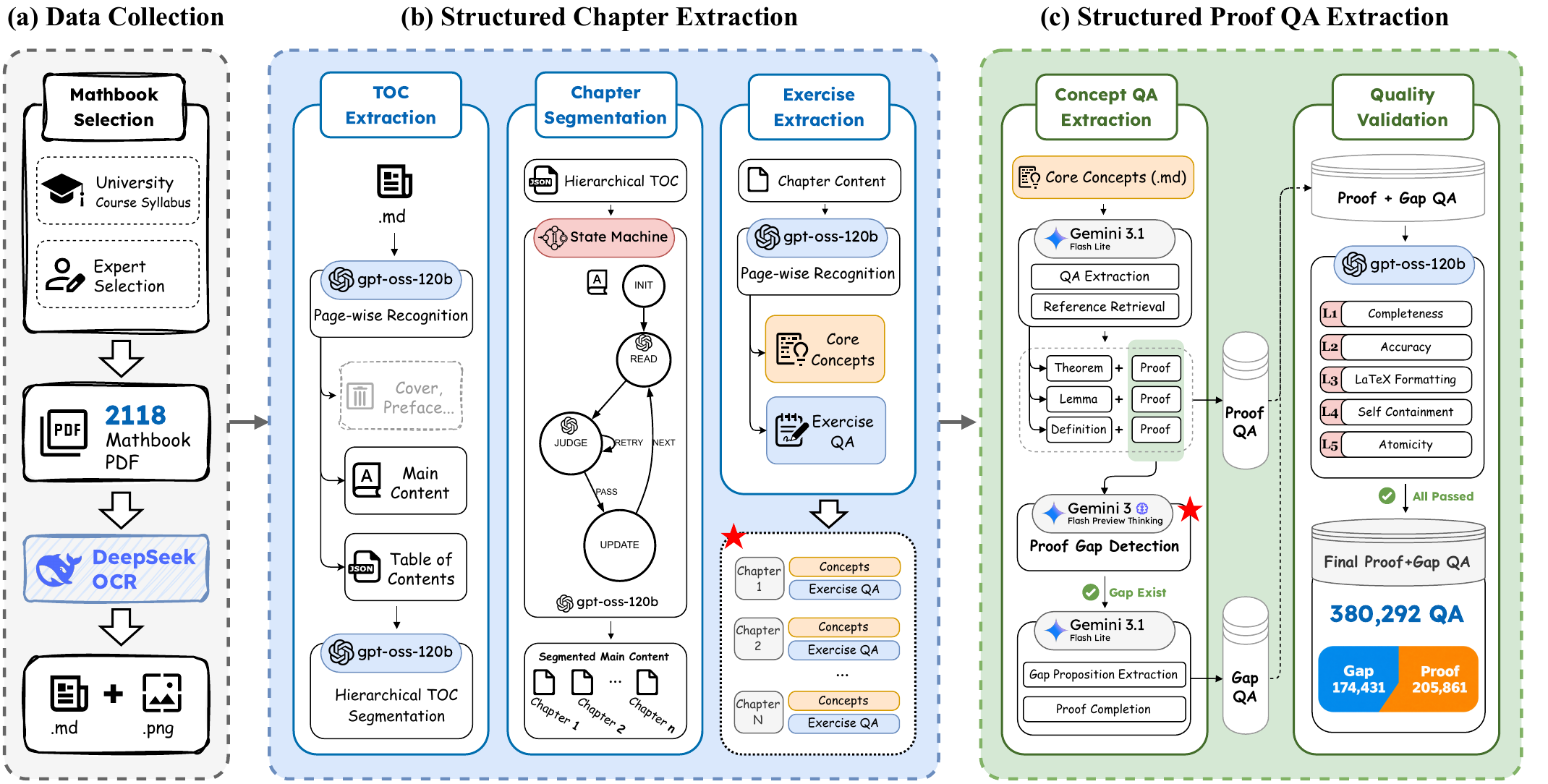}
\caption{Overview of TOME-Math construction, from structured textbook parsing and chapter recovery to Proof QA extraction, Gap QA generation, and quality-controlled corpus assembly (Section~\ref{sec:construction}).}

\label{fig:method}
\vspace{-10pt}
\end{figure}

\subsection{Data Construction}
\label{sec:construction}

\paragraph{Pipeline overview.}
We construct TOME-Math through a structure-preserving pipeline that converts raw textbook PDFs into two complementary forms of proof supervision: \textbf{Proof QA} derived from explicit proposition--proof units and \textbf{Gap QA} derived from missing proof obligations in Proof QAs. The source collection is selected against university course syllabi and reviewed by domain experts, yielding 2{,}118 input PDFs after canonicalization and deduplication (as shown in Section~\ref{sec:characterization}). The pipeline generated the full TOME-Math dataset of 380{,}292 QA pairs, comprising 174{,}431 Gap QA pairs and 205{,}861 Proof QA pairs.

\paragraph{Document parsing and structured chapter extraction.}
DeepSeek-OCR~\citep{wei2025deepseekocr} converts each PDF into multimodal Markdown together with page images. A page-wise \texttt{gpt-oss-120b}~\citep{openai2025gptoss} pass separates the table of contents and main body from covers, prefaces, and other front matter, then reconstructs a hierarchical table of contents. Conditioned on this hierarchy, a state machine reads the body in order, judges whether each page belongs to the current chapter, the next chapter, or contains an internal boundary, and retries ambiguous assignments before updating the chapter state. This produces an ordered set of chapter-level documents while preserving the textbook hierarchy. A second page-wise pass separates expository core concepts from textbook exercises within each chapter. The exercises are retained as an independent reservoir, but are excluded from the Proof QA generation path.

\paragraph{Proof QA Extraction and Gap QA construction.}
For each core concept chapter, we use \texttt{gemini-3.1-\allowbreak flash-\allowbreak lite-\allowbreak preview}~\citep{google2026gemini31flashlite} to extract theorem--proof, lemma--proof, and definition--proof units and retrieve the earlier propositions needed to interpret each proof. These source-grounded units form the Proof QA stream. We then use \texttt{gemini-3-flash-\allowbreak preview} with thinking enabled~\citep{google2025gemini3flash} to inspect each proof for \textbf{gaps} (\textit{i.e.}, implicit or missing reasoning steps). When a substantive gap is detected, \texttt{gemini-3.1-flash-\allowbreak lite-\allowbreak preview}~\citep{google2026gemini31flashlite} rewrites the missing obligation as a self-contained proposition and completes its proof, producing the Gap QA stream. This structured construction preserves explicit textbook proofs while adding targeted supervision for intermediate reasoning that the source leaves unstated.

\paragraph{Validation and quality control.}
Quality control is applied at both the structural and semantic levels. The structural checks validate the table-of-contents hierarchy, ambiguous page boundaries, question continuity across pages, question--answer alignment, and standalone completeness. For proposition extraction, a source-conditioned \texttt{gpt-oss-120b} judge audits candidates along five axes: completeness, accuracy, \LaTeX{} formatting, self-containment, and atomicity. Accuracy is treated as critical: invented proof steps, altered mathematical relations, or other content not supported by the source are failure cases rather than stylistic defects. Full module logic, intermediate artifacts, model-allocation details, and the judging rubric are given in Appendix~\ref{app:construction-details}.



\subsection{Characterization}
\label{sec:characterization}

\begin{figure}[t]
    \centering
    \begin{minipage}[b]{0.6\linewidth}
        \centering
        \includegraphics[width=\linewidth]{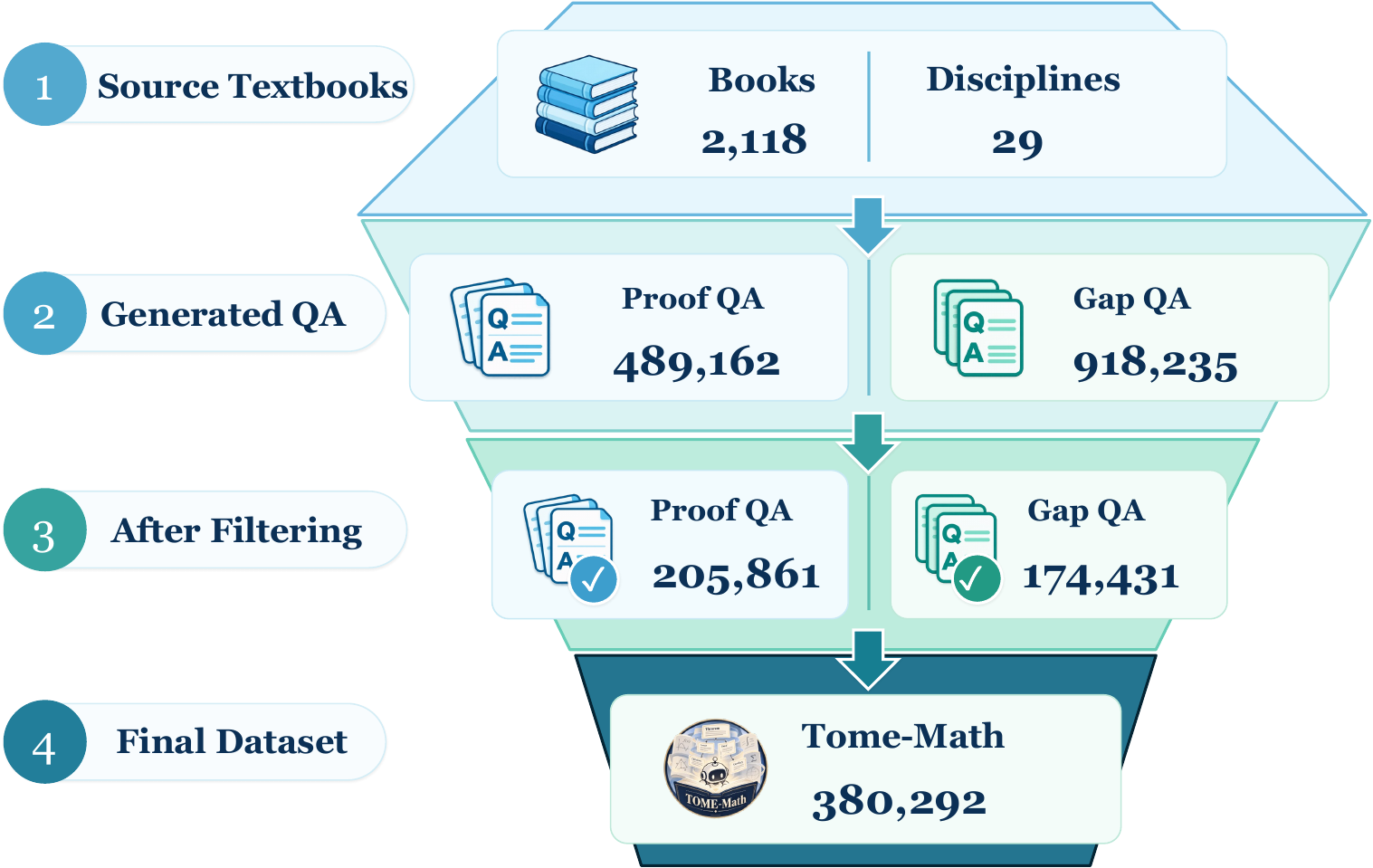}
        \vspace{-3pt}
        \small\textbf{(a)} Dataset construction and filtering.
    \end{minipage}%
    \hfill
    \begin{minipage}[b]{0.36\linewidth}
        \centering
        \includegraphics[width=\linewidth]{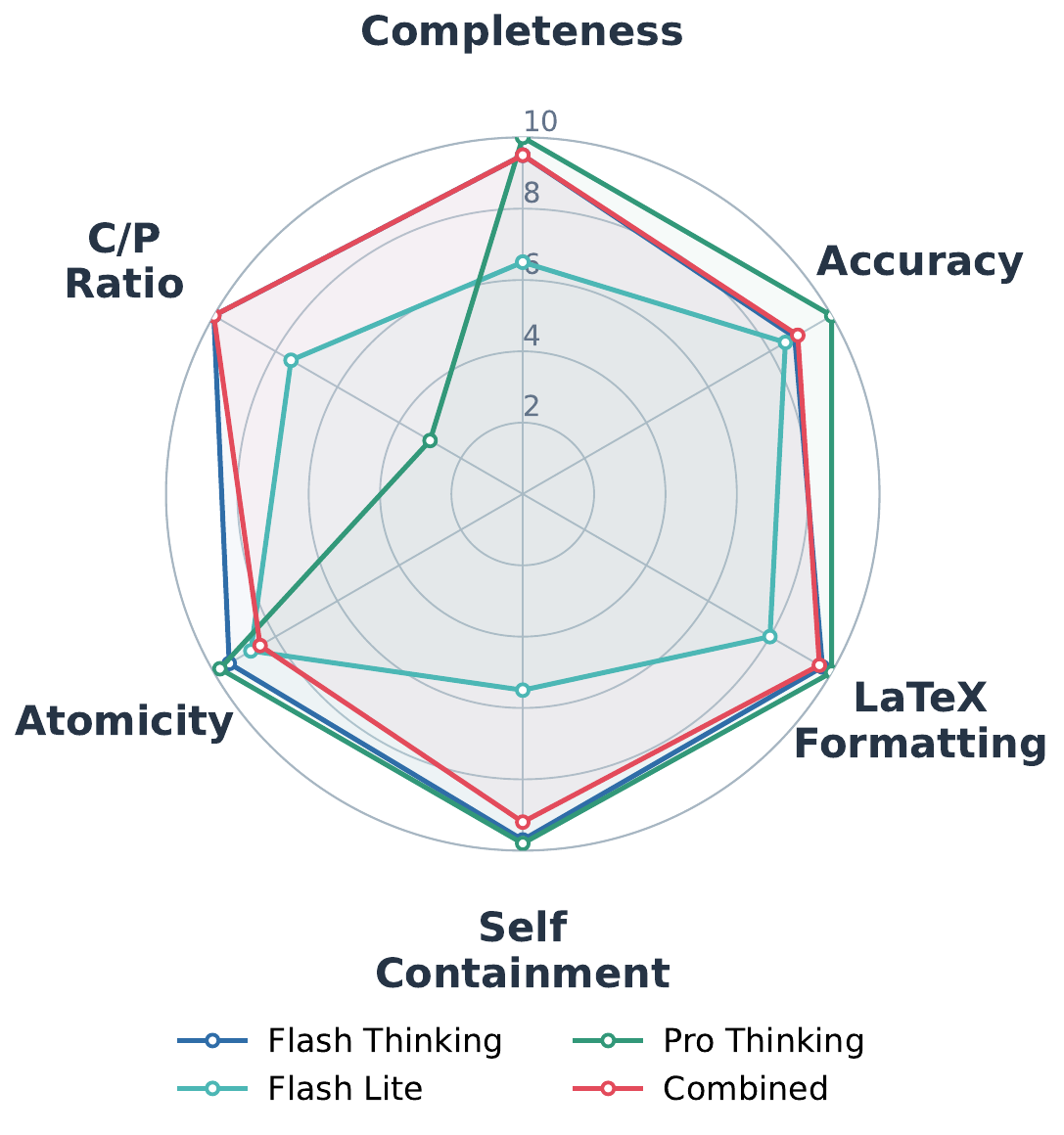}
        \small\textbf{(b)} Model-quality comparison.
    \end{minipage}
    \caption{TOME-Math data composition and quality characterization. (L):
    dataset scale from source textbooks through filtering to the final corpus.
    (R): model configurations compared across the quality-control criteria.}
    \label{fig:data-reserve-quality}
\end{figure}

TOME-Math comprises 380{,}292 QA records from 2{,}118 canonical textbooks
across 29 subject categories, summarized in Figure~\ref{fig:data-reserve-quality}.
Appendix~\ref{sec:characterization-appendix} provides the detailed corpus
characterization, including subject and language coverage, task composition,
structural complexity, confidence dynamics, quality, and provenance.

\paragraph{Generation Composition.}
\label{sec:char-scale}
The corpus combines a Proof QA stream of 205{,}861 records and a Gap QA stream of 174{,}431 records, drawing on disjoint model assignments (Appendix~\ref{sec:characterization-appendix}). \texttt{gemini-3-flash-preview \allowbreak -thinking}~\citep{google2025gemini3flash} produces 88.9\% of Proof QA records, \texttt{gpt-5-mini}~\citep{openai2025gpt5mini} produces 53.7\% of Gap QA records, and the Gap QA stream retains 174{,}431 of 918{,}235 generated candidates, a yield of 19.00\%.

\paragraph{Model-quality trade-off.} 
We evaluate model-quality trade-off along six dimensions, and combine \texttt{gemini-3\allowbreak-flash-preview-thinking} (Flash Thinking) for primary extraction and auditing with \texttt{gemini-3.1\allowbreak-flash-lite-preview} (Flash Lite) for reference-semantic parsing, balancing quality and cost. See more details in Appendix~\ref{app:model_quality}

\paragraph{Provenance and Evaluation Units.}
\label{sec:char-provenance}
Each record carries book, chapter, and source location metadata, with proposition identifiers and resolved references for Proof QA records. We define two evaluation scopes at the book level, held out exercises for item level generalization and held out books for source level generalization, both respecting duplicate groups identified by global content hash.

\section{Evaluating TOME-Math}

\label{sec:evaluating-tome}
\subsection{Experimental Setup}
\paragraph{Base Model.}
We use Qwen3-30B-A3B-Base~\citep{yang2025qwen3} as the backbone for all
experiments.  All SFT variants start from the same checkpoint, and the
unmodified checkpoint serves as the base-model baseline.

\paragraph{Evaluation.}
To obtain a sensitive view of how structured textbook supervision affects
learning, we focus on the six core subjects with the largest usable textbook
collections in our corpus: abstract algebra, differential geometry, discrete
mathematics, mathematical analysis, probability, and topology.  We construct
\textsc{\textbf{TOME-LongForm}} from their textbook-authored exercises: 300 proof-heavy
problems from held-out books, balanced across subjects and the earlier and
later halves of each book.  To enforce book-level separation, we reserve 64
source books for evaluation and exclude all 15{,}661 QA records associated with
those books from the complete 380{,}292-record TOME-Math corpus.  This leaves
364{,}631 examples in the actual training split.  The evaluation problems are
selected separately from the native-exercise reservoir of the held-out books.
We sample 16 solutions per problem and score them
with a reference-guided step verifier; each solution is reviewed eight times
by \texttt{gpt-5.6-luna}~\citep{openai2026gpt56luna} and is accepted only if no review identifies a
substantive error.
For complementary coverage beyond long-form proofs, \textsc{\textbf{TOME-Objective}}
contains 145 multidisciplinary objective problems scored by final-answer
equivalence.  We additionally evaluate on \textbf{MiniF2F}~\citep{zheng2021minif2f},
\textbf{ProofNet}~\citep{azerbayev2023proofnet}, \textbf{MATH-500}~\citep{hendrycks2021math}, and
\textbf{AIME2026}~\citep{dekoninck2026matharena}.
Full construction and
scoring details are given in Appendix~\ref{app:evaluation-protocol}.

\subsection{Full-Corpus Comparison}
\label{sec:main-results}

We first present the overall effectiveness of the final TOME-Math dataset,
before isolating its structural components in
Section~\ref{sec:understanding-textbook-supervision}.
Table~\ref{tab:main-hard-proof} compares the resulting 364{,}631-example
TOME-Math training split with three public mathematics SFT corpora at their
available scales.  All
fine-tuned models start from Qwen3-30B-A3B-Base and use the same fine-tuning
setup, differing only in their supervision corpus; each model is trained for
one full epoch over its respective corpus, and the unmodified checkpoint is
included as a reference.  Complete training configurations are provided in
Appendix~\ref{app:full-corpus-training}.  


\begin{table}[t]
\centering
\small
\setlength{\tabcolsep}{4pt}
\setlength{\abovecaptionskip}{6pt}
\setlength{\belowcaptionskip}{6pt}
\caption{\textbf{Full-corpus comparison of models independently fine-tuned on each listed SFT corpus.}
Evaluation covers \textsc{TOME-LongForm} (\textsc{TOME-LF.}),
\textsc{TOME-Objective}(\textsc{TOME-Obj.}), MiniF2F, ProofNet, MATH-500, and AIME 2026. Scores are
reported as percentages; Avg. is the unweighted mean of the six reported scores.
Best and second-best results in each column are shown in
\textbf{bold} and \underline{underlined}, respectively.}
\label{tab:main-hard-proof}
\resizebox{\textwidth}{!}{%
\begin{tabular}{lccccccc}
\toprule
SFT supervision corpus & \textsc{TOME-LF} & \textsc{TOME-Obj.} & MiniF2F & ProofNet & MATH-500 & AIME-26 & Avg. \\
\midrule
\rowcolor{tbGrayLight}
None (Qwen3-30B-A3B-Base) & 13.25 & 32.93 & 34.45 & 17.14 & 68.80 & 8.44 & 29.17 \\
\addlinespace[2pt]
OpenMath1 (1.82M)   & 20.16 & 33.86 & 22.87 & 19.25 & 70.40 & 8.21 & 29.13 \\
MetaMath (395k)     & 28.73 & 34.66 & 40.70 & 22.55 & 71.20 & 9.98 & 34.64 \\
MathInstruct (262k) & \underline{35.87} & \underline{40.69} & \underline{52.05} & \underline{32.93} & \underline{73.05} & \textbf{11.98} & \underline{41.10} \\
\rowcolor{tbTealLight}
TOME-Math (ours; 364.6k) & \textbf{43.86} & \textbf{43.96} & \textbf{57.43} & \textbf{42.98} & \textbf{76.55} & \underline{10.19} & \textbf{45.83} \\
\bottomrule
\end{tabular}%
}
\end{table}

\paragraph{Aggregate performance.}
TOME-Math achieves the best result on five of the six benchmarks and ranks
second on AIME 2026.  Its largest margins over the strongest public baseline
appear on the proof-oriented \textsc{TOME-LongForm} and ProofNet evaluations,
while its advantages on \textsc{TOME-Objective}, MiniF2F, and MATH-500 show
that the benefit extends beyond its own evaluation distribution and is not
confined to long-form proof generation.  Averaged uniformly across the six
reported scores, TOME-Math improves the base-model mean from 29.17\% to
45.83\%, a gain of 16.66 percentage points (16.7 when rounded to one decimal).

\paragraph{Effect of supervision corpus.}
The ranking is not explained by corpus size alone.  OpenMath1 contains 1.82M
examples, approximately $4.8\times$ the size of TOME-Math---yet TOME-Math
outperforms it on all six benchmarks.  It also exceeds the similarly sized
MetaMath corpus throughout and improves consistently over the pretrained
backbone.  These results demonstrate the overall utility of TOME-Math and
suggest that the source and form of supervision matter in addition to raw
example count.  Because the corpora differ in several respects, the comparison
does not by itself isolate textbook structure as the causal factor; we examine
that structure through controlled retrieval and training studies in
Section~\ref{sec:understanding-textbook-supervision}.

\section{Leveraging Textbook Structure}
\label{sec:understanding-textbook-supervision}

Beyond providing a large source of proof supervision, TOME-Math exposes structural signals---chapter progression, proposition dependencies, and source provenance---that can be used explicitly at inference time and during training. This section investigates when these signals improve mathematical reasoning. We first evaluate structure-guided retrieval without updating model parameters (Section~\ref{sec:dependency-grounded-proof-generation}). We then study whether textbook order provides an effective curriculum under controlled budgets (Section~\ref{sec:curriculum-scaling}), and examine source breadth together with transfer to held-out books from other mathematical fields (Section~\ref{sec:breadth-density}). Finally, we examine how these structural choices contribute to the full-corpus TOME-Math training recipe (Section~\ref{sec:full-corpus-ablation}).

\subsection{Training-Free Structure-Guided Retrieval}
\label{sec:dependency-grounded-proof-generation}

\paragraph{Question and design.}
Can textbook structure identify useful evidence for mathematical problem solving?
We test five paired context policies on all 300 \textsc{TOME-LongForm}
problems using a frozen Qwen3-30B-A3B-Base model:
\textbf{(i)} \textsc{Closed} uses no external context;
\textbf{(ii)} \textsc{Random Same-Book} samples uniformly from the exercise's
source book; \textbf{(iii)} \textsc{Local Chapter} draws only from the exercise
chapter and its immediate predecessors;
\textbf{(iv)} \textsc{TOME (Props Only)} retrieves proposition--proof records
using chapter position and recovered dependencies; and
\textbf{(v)} \textsc{TOME (Props + Gap)} extends the same policy to include
Gap QA.

All context-augmented conditions receive the same context-token budget.  They
also share the same backbone, prompt, decoding configuration, and
reference-guided verifier; the target exercise and material overlapping its
reference answer are removed from every retrieval pool.  The resulting
comparison asks how effectively each policy selects a fixed amount of
same-source evidence, rather than whether additional test-time tokens alone
help.  Appendix~\ref{app:dependency-context-protocol} gives the complete
retrieval, leakage-control, generation, and scoring protocol.

\begin{table}[t]
    \centering
    \small
    \setlength{\tabcolsep}{4.5pt}
    \setlength{\abovecaptionskip}{6pt}
    \setlength{\belowcaptionskip}{6pt}
    \caption{\textbf{Training-free structure-guided retrieval on
    \textsc{TOME-LongForm}.} Scores are pass@1 (\%) on the same 300 problems.
    Avg. is the mean over six equally sized subjects; the remaining columns
    report individual subjects. Best results are bolded. See
    Section~\ref{sec:dependency-grounded-proof-generation} for details.}
    \label{tab:dependency-grounded-generation}
    \resizebox{\linewidth}{!}{%
    \begin{tabular}{lc@{\hspace{12pt}}cccccc}
        \toprule
        Context & Avg. & Abs.\ Alg. & Diff.\ Geom. & Discrete & Math.\ Anal. & Prob. & Topology \\
        \midrule
        \textsc{Closed}             & 13.25          & 8.50           & 5.25           & 13.25          & 25.75          & 20.00          & 6.75           \\
        \textsc{Random Same-Book}   & 14.00          & 9.75           & 6.50           & 12.50          & 24.75          & 20.50          & 10.00          \\
        \textsc{Local Chapter}      & 14.71          & 10.50          & 6.75           & 14.50          & 25.50          & 20.00          & 11.00          \\
        \textsc{TOME (Props Only)}  & 15.54          & 11.00          & 8.50           & 15.00          & 27.25          & 20.50          & 11.00          \\
        \rowcolor{tbTealLight}
        \textsc{TOME (Props + Gap)} & \textbf{17.38} & \textbf{12.75} & \textbf{10.00} & \textbf{15.25} & \textbf{31.75} & \textbf{21.75} & \textbf{12.75} \\
        \bottomrule
    \end{tabular}}
\end{table}

\paragraph{Results.}
The results show a clear progression as increasingly structured evidence is
introduced.  Starting from 13.25\% pass@1 in the \textsc{Closed} setting,
\textsc{Random Same-Book} and \textsc{Local Chapter} reach 14.00\% and
14.71\%, respectively.  Retrieving proposition--proof records with textbook
structure further improves performance to 15.54\%, while augmenting this
inventory with Gap QA yields the best overall score of 17.38\%, a 4.13-point
gain over closed-book generation.  In particular, Gap QA contributes an
additional 1.84 points over proposition-only retrieval, showing that explicit
local proof obligations provide complementary evidence beyond textbook
propositions.  Among the four context-augmented conditions, the shared token
budget shows that their performance differences arise from evidence selection
rather than context quantity.

The advantage of the complete \textsc{TOME (Props + Gap)} policy is consistent
across all six subjects, with improvements of 1.75--6.00 points over
\textsc{Closed}.  Mathematical analysis and topology show the largest gains
(+6.00 points each), while every other subject also improves.  By contrast,
random same-book or local-chapter context occasionally leaves performance
unchanged or lowers it, indicating that access to nearby textbook material
alone is not sufficient.  Together, these results establish the retained
textbook structure as an effective inference-time evidence index and show that
proposition context and explicit proof obligations work best in combination.

\subsection{Curriculum Learning Across Subjects and Data Scales}
\label{sec:curriculum-scaling}

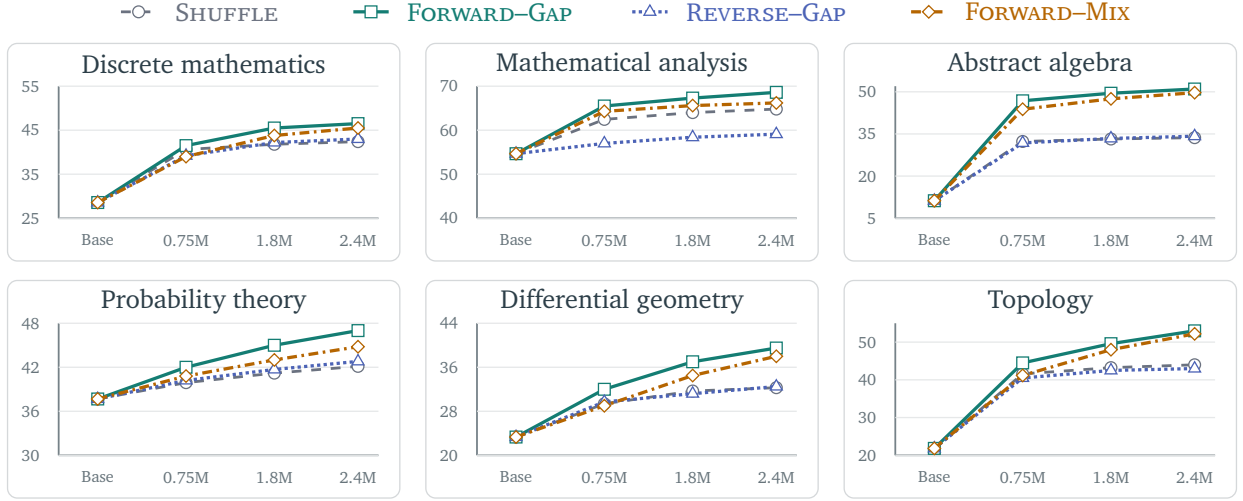
\begin{figure}[!t]
    \centering
    \resizebox{0.98\linewidth}{!}{%
    \begin{tikzpicture}[
        x=1cm,
        y=1cm,
        axis/.style={draw=tbInk!65, line width=0.55pt},
        grid/.style={draw=tbInk!12, line width=0.4pt},
        randomline/.style={draw=schedShuffle, dashed, line width=0.85pt},
        forwardline/.style={draw=schedForwardGap, solid, line width=1.0pt},
        reverseline/.style={draw=schedReverseGap, densely dotted, line width=1.0pt},
        mixedline/.style={draw=schedForwardMix, dash pattern=on 3pt off 1.4pt on 0.8pt off 1.4pt,
            line width=1.0pt},
        randompoint/.style={circle, draw=schedShuffle, fill=white,
            minimum size=3.6pt, inner sep=0pt},
        forwardpoint/.style={rectangle, draw=schedForwardGap, fill=white,
            minimum size=3.6pt, inner sep=0pt},
        reversepoint/.style={regular polygon, regular polygon sides=3,
            draw=schedReverseGap, fill=white, minimum size=4.2pt, inner sep=0pt},
        mixedpoint/.style={diamond, draw=schedForwardMix, fill=white,
            minimum size=4.0pt, inner sep=0pt},
        paneltitle/.style={font=\footnotesize, text=tbInk},
        legendlabel/.style={font=\footnotesize},
        tick/.style={font=\tiny, text=tbInk!72}
    ]
        \foreach \xoff/\yoff/\title in {
            0/2.55/Discrete mathematics,
            4.50/2.55/Mathematical analysis,
            9.00/2.55/Abstract algebra,
            0/0/Probability theory,
            4.50/0/Differential geometry,
            9.00/0/Topology} {
            \path[draw=tbInk!20, fill=white, rounded corners=3pt]
                (\xoff+0.03,\yoff+0.08) rectangle (\xoff+4.25,\yoff+2.42);
            \node[paneltitle] at (\xoff+2.14,\yoff+2.20) {\title};

            \draw[axis] (\xoff+0.58,\yoff+0.54) -- (\xoff+0.58,\yoff+1.96);
            \draw[axis] (\xoff+0.58,\yoff+0.54) -- (\xoff+4.02,\yoff+0.54);
            \foreach \gx/\lab in {1.00/Base,1.95/0.75M,2.90/1.8M,3.80/2.4M}
                \node[tick, anchor=north] at (\xoff+\gx,\yoff+0.49) {\lab};
        }

        \newcommand{\currpanelticks}[5]{%
            \foreach \ytick in {#5} {%
                \pgfmathsetmacro{\ytickpos}{#2+0.54+1.42*(\ytick-#3)/(#4-#3)}
                \draw[grid] (#1+0.58,\ytickpos) -- (#1+4.02,\ytickpos);
                \node[tick,anchor=east] at (#1+0.50,\ytickpos) {\ytick};
            }
        }
        \currpanelticks{0}{2.55}{25}{55}{25,35,45,55}
        \currpanelticks{4.50}{2.55}{40}{70}{40,50,60,70}
        \currpanelticks{9.00}{2.55}{5}{52}{5,20,35,50}
        \currpanelticks{0}{0}{30}{48}{30,36,42,48}
        \currpanelticks{4.50}{0}{20}{44}{20,28,36,44}
        \currpanelticks{9.00}{0}{20}{55}{20,30,40,50}

        \newcommand{\currplotfour}[6]{%
            \draw[#1]
                (0.42,{(#3-\currmin)*\currscale}) --
                (1.37,{(#4-\currmin)*\currscale}) --
                (2.32,{(#5-\currmin)*\currscale}) --
                (3.22,{(#6-\currmin)*\currscale});
            \foreach \px/\py in {0.42/#3,1.37/#4,2.32/#5,3.22/#6}
                \node[#2] at (\px,{(\py-\currmin)*\currscale}) {};
        }
        \begin{scope}[shift={(0.58,3.09)}]
            \def\currmin{25}\pgfmathsetmacro{\currscale}{1.42/(55-25)}
            \currplotfour{randomline}{randompoint}{28.5664}{40.6174}{41.8000}{42.4000}
            \currplotfour{forwardline}{forwardpoint}{28.5664}{41.5000}{45.5000}{46.5000}
            \currplotfour{reverseline}{reversepoint}{28.5664}{39.2000}{42.2000}{43.0000}
            \currplotfour{mixedline}{mixedpoint}{28.5664}{39.0000}{43.8000}{45.5000}
        \end{scope}

        \begin{scope}[shift={(5.08,3.09)}]
            \def\currmin{40}\pgfmathsetmacro{\currscale}{1.42/(70-40)}
            \currplotfour{randomline}{randompoint}{54.6611}{62.4595}{64.0000}{64.8000}
            \currplotfour{forwardline}{forwardpoint}{54.6611}{65.5000}{67.3000}{68.6000}
            \currplotfour{reverseline}{reversepoint}{54.6611}{57.0000}{58.4000}{59.1000}
            \currplotfour{mixedline}{mixedpoint}{54.6611}{64.2850}{65.6000}{66.2000}
        \end{scope}

        \begin{scope}[shift={(9.58,3.09)}]
            \def\currmin{5}\pgfmathsetmacro{\currscale}{1.42/(52-5)}
            \currplotfour{randomline}{randompoint}{11.2276}{32.3213}{33.2000}{33.6756}
            \currplotfour{forwardline}{forwardpoint}{11.2276}{46.8000}{49.5000}{51.0000}
            \currplotfour{reverseline}{reversepoint}{11.2276}{31.8000}{33.4000}{34.2000}
            \currplotfour{mixedline}{mixedpoint}{11.2276}{43.8000}{47.5000}{49.7000}
        \end{scope}

        \begin{scope}[shift={(0.58,0.54)}]
            \def\currmin{30}\pgfmathsetmacro{\currscale}{1.42/(48-30)}
            \currplotfour{randomline}{randompoint}{37.6745}{39.9000}{41.2000}{42.1234}
            \currplotfour{forwardline}{forwardpoint}{37.6745}{42.0000}{45.0000}{47.0000}
            \currplotfour{reverseline}{reversepoint}{37.6745}{40.2000}{41.7000}{42.8000}
            \currplotfour{mixedline}{mixedpoint}{37.6745}{40.8000}{43.0000}{44.8000}
        \end{scope}

        \begin{scope}[shift={(5.08,0.54)}]
            \def\currmin{20}\pgfmathsetmacro{\currscale}{1.42/(44-20)}
            \currplotfour{randomline}{randompoint}{23.3159}{29.3676}{31.6814}{32.2874}
            \currplotfour{forwardline}{forwardpoint}{23.3159}{32.0000}{37.0000}{39.5000}
            \currplotfour{reverseline}{reversepoint}{23.3159}{29.7000}{31.2000}{32.5000}
            \currplotfour{mixedline}{mixedpoint}{23.3159}{29.0000}{34.5000}{38.0000}
        \end{scope}

        \begin{scope}[shift={(9.58,0.54)}]
            \def\currmin{20}\pgfmathsetmacro{\currscale}{1.42/(55-20)}
            \currplotfour{randomline}{randompoint}{21.8243}{41.5799}{43.2387}{44.0000}
            \currplotfour{forwardline}{forwardpoint}{21.8243}{44.5000}{49.6000}{53.0000}
            \currplotfour{reverseline}{reversepoint}{21.8243}{40.5000}{42.5000}{43.0000}
            \currplotfour{mixedline}{mixedpoint}{21.8243}{41.1756}{48.0000}{52.2000}
        \end{scope}

        \begin{scope}[xshift=0.4cm]
            \draw[randomline] (0.85,5.31) -- (1.23,5.31);
            \node[randompoint] at (1.04,5.31) {};
            \node[legendlabel,anchor=west] at (1.31,5.31) {\schedshuffle};
            \draw[forwardline] (3.35,5.31) -- (3.73,5.31);
            \node[forwardpoint] at (3.54,5.31) {};
            \node[legendlabel,anchor=west] at (3.81,5.31) {\schedforwardgap};
            \draw[reverseline] (6.35,5.31) -- (6.73,5.31);
            \node[reversepoint] at (6.54,5.31) {};
            \node[legendlabel,anchor=west] at (6.81,5.31) {\schedreversegap};
            \draw[mixedline] (9.35,5.31) -- (9.73,5.31);
            \node[mixedpoint] at (9.54,5.31) {};
            \node[legendlabel,anchor=west] at (9.81,5.31) {\schedforwardmix};
        \end{scope}
    \end{tikzpicture}}
    \caption{\textbf{Curriculum performance across subjects and target-token budgets.}
    Pass@16 is the fraction of problems with at least one verifier-accepted
    solution among 16 samples. Within each subject--scale cell, all schedules
    use identical examples and budgets; Base denotes the shared pretrained
    checkpoint. See Section~\ref{sec:curriculum-scaling} for details.}
    \label{fig:curriculum-scaling}
\end{figure}

The retrieval study shows that textbook structure can guide the selection of
useful evidence at inference time.  We next ask whether the same organization
can also improve learning when it is used to order supervision, and whether
any benefit persists across subjects and data scales.

\paragraph{Setup and comparisons.}
We evaluate textbook-aligned curricula across six subjects and three
per-subject target-token budgets: 0.75M, 1.8M, and 2.4M. Within each
subject--scale cell, four schedules reorder the same frozen QA set:
\schedshuffle, \schedforwardgap, \schedreversegap, and \schedforwardmix. Here,
Gap QA makes a locally omitted proof obligation explicit. \schedforwardgap{}
versus \schedreversegap{} isolates chapter direction under a fixed Gap-first
policy, while \schedforwardgap{} versus \schedforwardmix{} isolates Gap-first
ordering under a fixed forward progression; comparison with \schedshuffle{}
measures their combined effect. All schedules share the same examples, token
budget, initialization, and optimization settings, and are evaluated on the
same held-out-book exercises.
Appendix~\ref{app:curriculum-scaling-protocol} details the normalized chapter
stages, schedule construction, nested data scales, and order-preserving
training controls.

\paragraph{Overall effect of scale and curriculum.}
Figure~\ref{fig:curriculum-scaling} shows that every fine-tuned condition
outperforms the shared base checkpoint in all six subjects, even at 0.75M
tokens, and that every schedule improves with scale in each subject.
Across the 0.75M, 1.8M, and 2.4M budgets, macro-average pass@16 rises from
41.04\% to 42.52\% and 43.21\% for \schedshuffle{}, and from 45.38\% to
48.98\% and 50.93\% for \schedforwardgap{}. Thus, both supervision scale and
structured ordering improve solution coverage.

\paragraph{Effect of chapter direction and Gap-first ordering.}
\schedforwardgap{} is strongest in all 18 subject--scale cells, and its
macro-average margin over \schedshuffle{} widens from 4.34 points at 0.75M
tokens to 6.46 at 1.8M and 7.72 at 2.4M. Controlled comparisons attribute most
of this benefit to chapter direction: relative to \schedreversegap, the forward
schedule gains 5.65, 7.41, and 8.50 points across the three scales. Holding
forward progression fixed, Gap-first ordering adds a smaller but consistent
2.37, 1.91, and 1.53 points over \schedforwardmix{}. Thus, forward progression
is the primary contributor, with Gap-first ordering providing a complementary
gain. The same ordering across all 18 subject--scale cells shows that these
benefits persist across mathematical domains and supervision budgets.


\subsection{Source Breadth and Cross-Subject Transfer}
\label{sec:breadth-density}

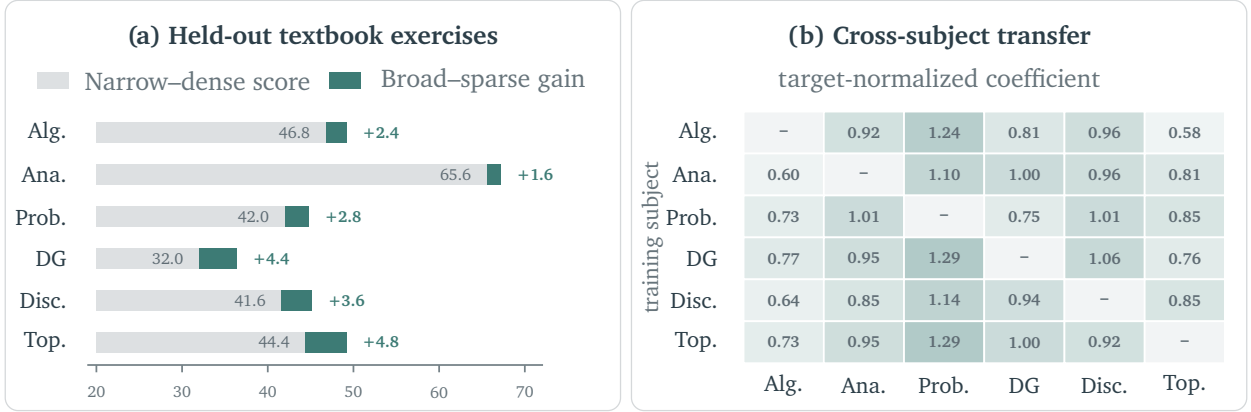
\begin{figure}[!t]
    \centering
    \resizebox{0.99\linewidth}{!}{%
    \begin{tikzpicture}[
        x=1cm,
        y=1cm,
        font=\scriptsize,
        frame/.style={draw=tbInk!20, fill=white, rounded corners=4pt},
        paneltitle/.style={font=\bfseries\footnotesize, text=tbInk},
        annotation/.style={font=\footnotesize, text=tbInk!72},
        note/.style={font=\tiny, text=tbInk!72},
        barbase/.style={draw=none, fill=tbInk!16},
        bargain/.style={draw=none, fill=tbTeal},
        axis/.style={draw=tbInk!65, line width=0.55pt}
    ]
        \path[frame] (0.55,0.50) rectangle (7.30,5.00);
        \node[paneltitle] at (3.925,4.60)
            {(a) Held-out textbook exercises};

        \begin{scope}[xshift=0.2cm]
            \path[barbase] (0.70,4.03) rectangle (1.05,4.21);
            \node[anchor=west,annotation] at (1.10,4.12) {Narrow--dense score};
            \path[bargain] (3.90,4.03) rectangle (4.25,4.21);
            \node[anchor=west,annotation] at (4.35,4.12) {Broad--sparse gain};
        \end{scope}

        \draw[axis] (1.45,1.00) -- (6.45,1.00);
        \foreach \xx/\lab in {
            1.55/20,2.49/30,3.43/40,4.37/50,5.31/60,6.25/70} {
            \draw[axis] (\xx,1.00) -- (\xx,0.92);
            \node[note,anchor=north] at (\xx,0.89) {\lab};
        }
        \newcommand{\breadthpair}[4]{%
            \pgfmathsetmacro{\rowy}{3.55-0.46*#1}%
            \pgfmathsetmacro{\xn}{1.55+(#3-20)*0.094}%
            \pgfmathsetmacro{\xb}{1.55+(#3+#4-20)*0.094}%
            \node[anchor=east,font=\scriptsize,text=tbInk] at (1.35,\rowy) {#2};
            \path[barbase] (1.55,\rowy-0.11) rectangle (\xn,\rowy+0.11);
            \path[bargain] (\xn,\rowy-0.11) rectangle (\xb,\rowy+0.11);
            \node[anchor=east,font=\tiny,text=tbInk!76]
                at (\xn-0.05,\rowy) {#3};
            \node[anchor=west,font=\tiny\bfseries,text=tbTeal]
                at (\xb+0.05,\rowy) {+#4};
        }
        \breadthpair{0}{Alg.}{46.8}{2.4}
        \breadthpair{1}{Ana.}{65.6}{1.6}
        \breadthpair{2}{Prob.}{42.0}{2.8}
        \breadthpair{3}{DG}{32.0}{4.4}
        \breadthpair{4}{Disc.}{41.6}{3.6}
        \breadthpair{5}{Top.}{44.4}{4.8}

        \path[frame] (7.42,0.50) rectangle (14.17,5.00);
        \node[paneltitle] at (10.795,4.60)
            {(b) Cross-subject transfer};
        \node[annotation] at (10.795,4.12)
            {target-normalized coefficient};

        \newcommand{\transfercell}[4]{%
            \pgfmathsetmacro{\cellx}{8.65+0.88*#1}%
            \pgfmathsetmacro{\celly}{3.78-0.46*#2}%
            \path[draw=white,line width=0.55pt,fill=#3]
                (\cellx,\celly) rectangle ++(0.88,-0.46);
            \node[font=\tiny\bfseries,text=tbInk!76] at
                (\cellx+0.44,\celly-0.23) {#4};
        }
        \newcommand{\diagcell}[2]{%
            \transfercell{#1}{#2}{tbGrayLight}{--}%
        }

        \diagcell{0}{0}
        \transfercell{1}{0}{tbTeal!23}{0.92}
        \transfercell{2}{0}{tbTeal!37}{1.24}
        \transfercell{3}{0}{tbTeal!18}{0.81}
        \transfercell{4}{0}{tbTeal!25}{0.96}
        \transfercell{5}{0}{tbTeal!8}{0.58}

        \transfercell{0}{1}{tbTeal!9}{0.60}
        \diagcell{1}{1}
        \transfercell{2}{1}{tbTeal!31}{1.10}
        \transfercell{3}{1}{tbTeal!27}{1.00}
        \transfercell{4}{1}{tbTeal!25}{0.96}
        \transfercell{5}{1}{tbTeal!18}{0.81}

        \transfercell{0}{2}{tbTeal!14}{0.73}
        \transfercell{1}{2}{tbTeal!27}{1.01}
        \diagcell{2}{2}
        \transfercell{3}{2}{tbTeal!15}{0.75}
        \transfercell{4}{2}{tbTeal!27}{1.01}
        \transfercell{5}{2}{tbTeal!20}{0.85}

        \transfercell{0}{3}{tbTeal!16}{0.77}
        \transfercell{1}{3}{tbTeal!25}{0.95}
        \transfercell{2}{3}{tbTeal!40}{1.29}
        \diagcell{3}{3}
        \transfercell{4}{3}{tbTeal!29}{1.06}
        \transfercell{5}{3}{tbTeal!15}{0.76}

        \transfercell{0}{4}{tbTeal!11}{0.64}
        \transfercell{1}{4}{tbTeal!20}{0.85}
        \transfercell{2}{4}{tbTeal!33}{1.14}
        \transfercell{3}{4}{tbTeal!24}{0.94}
        \diagcell{4}{4}
        \transfercell{5}{4}{tbTeal!20}{0.85}

        \transfercell{0}{5}{tbTeal!14}{0.73}
        \transfercell{1}{5}{tbTeal!25}{0.95}
        \transfercell{2}{5}{tbTeal!40}{1.29}
        \transfercell{3}{5}{tbTeal!27}{1.00}
        \transfercell{4}{5}{tbTeal!23}{0.92}
        \diagcell{5}{5}

        \foreach \idx/\lab in {
            0/Alg.,
            1/Ana.,
            2/Prob.,
            3/DG,
            4/Disc.,
            5/Top.} {
            \pgfmathsetmacro{\labelx}{9.09+0.88*\idx}
            \node[font=\scriptsize,text=tbInk] at (\labelx,0.78) {\lab};
        }
        \foreach \idx/\lab in {
            0/Alg.,
            1/Ana.,
            2/Prob.,
            3/DG,
            4/Disc.,
            5/Top.} {
            \pgfmathsetmacro{\labely}{3.55-0.46*\idx}
            \node[font=\scriptsize,text=tbInk,anchor=east]
                at (8.49,\labely) {\lab};
        }
        \node[annotation,font=\scriptsize,rotate=90] at (7.65,2.40)
            {training subject};
    \end{tikzpicture}}
\caption{\textbf{Source breadth and cross-subject transfer.}
(a) Matched-budget pass@16 for 8-book narrow--dense training, with gains from
spreading the same budget across 16 books. (b) Target-normalized cross-subject
pass@16 for subject-specific checkpoints. Details are in Section~\ref{sec:breadth-density}.}
    
    \label{fig:source-breadth-transfer}
\end{figure}

Section~\ref{sec:curriculum-scaling} shows that performance improves as the
subject-specific training sets expand, but token volume and source count grow
together.  We therefore ask whether broader textbook coverage helps at a fixed
token budget and whether subject-specific supervision generalizes across
fields.  Figure~\ref{fig:source-breadth-transfer} summarizes these breadth and
transfer studies.

\paragraph{Source breadth.}
Figure~\ref{fig:source-breadth-transfer}(a) compares the 8-book
\textsc{Narrow--dense} condition with \textsc{Broad--sparse}, which distributes
the same 0.75M-token \schedforwardgap{} budget across 16 books while holding all
other settings fixed. \textsc{Broad--sparse} improves pass@16 in every subject
by 1.6--4.8 points (3.3 on average), with the largest gains in topology (+4.8),
differential geometry (+4.4), and discrete mathematics (+3.6). Thus, source
breadth is a distinct scaling axis: the same supervision budget is more
effective when spread across twice as many books. Together with the monotonic
gains in Section~\ref{sec:curriculum-scaling}, this result shows that textbook
supervision benefits from both greater volume and broader coverage.
Table~\ref{tab:curriculum-scale-composition} details the source pools and QA
counts.

\paragraph{Cross-subject generalization.}
Figure~\ref{fig:source-breadth-transfer}(b) evaluates each subject-specific
0.75M-token checkpoint on 50 exercises from every other field, normalized by
the target specialist's in-domain score,
$T_{s\rightarrow t}=P_{s\rightarrow t}/P_{t\rightarrow t}$. The macro-average
transfer coefficient is 0.92. Topology (0.98) and differential geometry (0.97)
have the strongest source-side means, and eight of the 30 ordered pairs exceed
1.0; both fields reach 1.29 on probability targets. This strong off-diagonal
performance shows that subject-specific supervision develops reusable proof
and reasoning patterns rather than benefits confined to its training field.
It also supports multidisciplinary training at full scale. Full results are in
Appendix~\ref{app:cross-subject-transfer-protocol}.

Taken together, the matched-budget breadth gains and cross-subject transfer
connect the controlled studies to the full TOME-Math corpus. Scaling to more
books adds source coverage without increasing the token budget, while scaling
across subjects supplies supervision that generalizes beyond its
domain. These complementary effects explain the effectiveness of the broad,
multidisciplinary corpus beyond example count alone.

\subsection{Full-Corpus Ablation of the TOME-Math Training Recipe}
\label{sec:full-corpus-ablation}

Finally, we ablate curriculum order, Gap-first scheduling, and Gap QA from the
full \schedforwardgap{} recipe and report pass@1 (\%) for all variants.
All variants use the book-held-out training split described in Section~\ref{sec:evaluating-tome}. 

\begin{table}[t]
    \centering
    \small
    \setlength{\tabcolsep}{4pt}
    \setlength{\abovecaptionskip}{6pt}
    \setlength{\belowcaptionskip}{6pt}
    \caption{\textbf{Full-corpus ablation of the TOME-Math training recipe.}
    See
    Section~\ref{sec:full-corpus-ablation} for details.}
    \label{tab:full-corpus-ablation}
    \begin{tabular*}{\linewidth}{@{\extracolsep{\fill}}lcccccc@{}}
        \toprule
        Variant & \textsc{TOME-LF} & \textsc{TOME-Obj.} & MiniF2F & ProofNet & MATH-500 & AIME-26 \\
        \midrule
        w/o Gap QA (\textsc{Forward})       & 38.72 & 38.91 & 50.73 & 36.21 & 73.84 & 9.12 \\
        w/o curriculum (\schedshuffle)         & 39.87 & 40.42 & 53.76 & 39.12 & 75.22 & 9.58 \\
        w/o gap-first (\schedforwardmix)     & 40.12 & 41.37 & 55.61 & 41.02 & 75.88 & 9.93 \\
        \rowcolor{tbTealLight}
        TOME-Math (\schedforwardgap)          & 43.86 & 43.96 & 57.43 & 42.98 & 76.55 & 10.19 \\
        \bottomrule
    \end{tabular*}
\end{table}

\paragraph{Component contributions.}
Table~\ref{tab:full-corpus-ablation} shows that every ablation lowers pass@1
across all six benchmarks. On the two TOME benchmarks, replacing Gap-first
with within-stage mixing costs 3.74 and 2.59 points, while global shuffling
costs another 0.25 and 0.95 points. Thus, Gap-first is the primary ordering
signal, complemented by forward chapter progression. Removing Gap QA causes
the largest drops on the four proof benchmarks (5.05--6.77 points), confirming
the value of explicit local proof obligations. The complete recipe also leads
on MATH-500 and AIME-26, extending the gains beyond proof-focused evaluations.

\paragraph{Structural synergy.}
The ablations reveal a mutually reinforcing relationship between TOME-Math's
corpus construction and training strategy. Gap QA introduces explicit proof
obligations, Gap-first scheduling prioritizes them within each stage, and
forward progression exploits the preserved textbook order. The consistent
gains across textbook-derived proof tasks, public proof benchmarks, and competition problems
establish TOME-Math's advantage as structured supervision beyond scale alone.

\section{Conclusion}
We introduced TOME-Math, a corpus of 380K proof records from 2{,}118 textbooks that preserves curricular and dependency structure. It improves performance across textbook-derived and public benchmarks, while controlled studies demonstrate the benefits of structure-guided retrieval, curriculum ordering, and broader source coverage. We hope TOME-Math inspires mathematical reasoning systems that learn not only from more data, but from better-organized knowledge.

\clearpage
\bibliography{iclr2027_conference}
\bibliographystyle{plainnat}

\newpage
\appendix
\startcontents[appendix]

\section*{Appendix Contents}
\begingroup
    \hypersetup{linkcolor=tbInk}
    \renewcommand{\contentsname}{}
    \titlecontents{section}
        [1.8em]
        {\normalsize\bfseries\addvspace{0.45em}}
        {\contentslabel{1.8em}}
        {}
        {\titlerule*[0.55em]{.}\contentspage}
    \titlecontents{subsection}
        [2.8em]
        {\small}
        {\contentslabel{2.8em}}
        {}
        {\titlerule*[0.55em]{.}\contentspage}
    \printcontents[appendix]{}{1}[2]{}
\endgroup
\clearpage

\section{TOME-Math Construction Details}
\label{app:construction-details}

This appendix expands the construction stages summarized in Section~\ref{sec:construction}. We distinguish the artifacts produced by the book-structure modules (M1--M3), exercise and answer processing (M4--M6), and the downstream proposition-extraction track.

\subsection{Intermediate representation}

DeepSeek-OCR produces one multimodal-Markdown representation of the complete book together with the extracted images. The automated modules materialize the following intermediate artifacts:

\begin{center}
    \small
    \begin{tabular}{p{0.27\linewidth}p{0.61\linewidth}}
        \hline
        \textbf{Artifact} & \textbf{Contents} \\
        \hline
        \texttt{toc.json} & Multilevel table of contents represented by title
        and heading depth. \\
        \texttt{main\_text.md} & Textbook body after removing front matter. \\
        \texttt{images/} & Merged image directory with normalized references
        from the Markdown files. \\
        \texttt{content/} & One Markdown file for every recovered chapter. \\
        \texttt{content\_pure/} & Chapter text after exercise segments are
        removed. \\
        \texttt{exercise/} & Chapter-indexed exercise text and formatted
        per-chapter JSONL files. \\
        \texttt{exercise\_index.jsonl} & Multilevel heading path for every
        exercise section. \\
        \texttt{answer.md} & A detached answer chapter when one is present. \\
        \hline
    \end{tabular}
\end{center}

\begin{figure}[t]
    \centering
    \resizebox{0.96\linewidth}{!}{%
    \begin{tikzpicture}[
        x=1cm,
        y=1cm,
        >=Latex,
        font=\small,
        rail/.style={draw=tbBlue!18, line width=5pt, line cap=round},
        flow/.style={->, draw=tbBlue, line width=0.9pt},
        stage/.style={circle, minimum size=9.5mm, inner sep=0pt,
            fill=tbBlue, text=white, font=\bfseries\small},
        title/.style={font=\bfseries\small, text=tbInk, align=center},
        proof/.style={draw=tbTeal, fill=tbTealLight, rounded corners=4pt,
            minimum width=3.7cm, minimum height=11mm, align=left,
            inner xsep=8pt, line width=0.8pt},
        eval/.style={draw=tbAmber, fill=tbAmberLight, rounded corners=4pt,
            minimum width=3.7cm, minimum height=11mm, align=left,
            inner xsep=8pt, line width=0.8pt}
    ]
        \node[circle, fill=tbGrayLight, draw=tbInk!45, minimum size=12mm,
            inner sep=0pt] (book) at (0.7,0) {};
        \draw[draw=tbInk, line width=0.75pt, rounded corners=0.5pt]
            ($(book.center)+(-0.34,0.22)$) -- ($(book.center)+(-0.03,0.13)$) --
            ($(book.center)+(-0.03,-0.24)$) -- ($(book.center)+(-0.34,-0.15)$) -- cycle;
        \draw[draw=tbInk, line width=0.75pt, rounded corners=0.5pt]
            ($(book.center)+(0.34,0.22)$) -- ($(book.center)+(0.03,0.13)$) --
            ($(book.center)+(0.03,-0.24)$) -- ($(book.center)+(0.34,-0.15)$) -- cycle;
        \node[title, below=5mm of book] {Source textbook};

        \node[stage] (parse) at (3.25,0) {1};
        \node[title, below=5mm of parse] {Parse pages};

        \node[stage] (hierarchy) at (6.35,0) {2};
        \node[title, below=5mm of hierarchy] {Recover hierarchy};

        \node[stage] (separate) at (9.45,0) {3};
        \node[title, below=5mm of separate] {Separate streams};

        \draw[rail] (book.east) -- (parse.west);
        \draw[rail] (parse.east) -- (hierarchy.west);
        \draw[rail] (hierarchy.east) -- (separate.west);
        \draw[flow] (book.east) -- (parse.west);
        \draw[flow] (parse.east) -- (hierarchy.west);
        \draw[flow] (hierarchy.east) -- (separate.west);

        \coordinate (fork) at (11.05,0);
        \node[proof] (proof) at (14.0,0.72)
            {\textbf{Expository stream}\\[-1pt]
             \scriptsize proof supervision};
        \node[eval] (evaluation) at (14.0,-0.72)
            {\textbf{Exercise stream}\\[-1pt]
             \scriptsize textbook evaluation};

        \draw[flow] (separate.east) -- (fork);
        \draw[->, draw=tbTeal, line width=0.95pt, rounded corners=6pt]
            (fork) |- (proof.west);
        \draw[->, draw=tbAmber, line width=0.95pt, rounded corners=6pt]
            (fork) |- (evaluation.west);
        \fill[tbTeal] ($(proof.west)+(0.12,0)$) circle (1.5pt);
        \fill[tbAmber] ($(evaluation.west)+(0.12,0)$) circle (1.5pt);
    \end{tikzpicture}}
    \caption{High-level construction flow.  Structural recovery separates
    expository material from native exercises before the two streams are used
    for proof supervision and textbook evaluation.}
    \label{fig:appendix-structure-pipeline}
\end{figure}
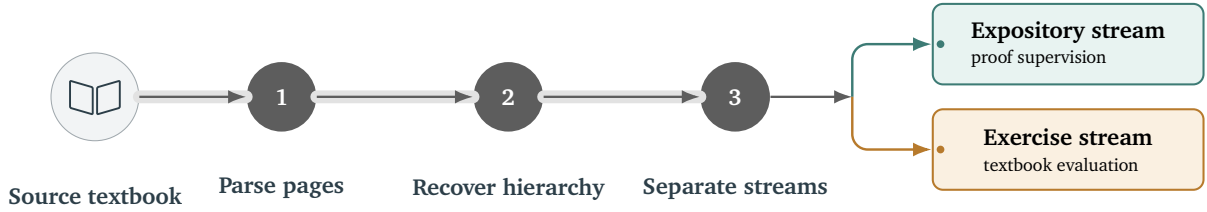

\subsection{M1--M2: Table-of-contents and chapter recovery}

\paragraph{M1: front matter and table of contents.}
Starting from the first parsed page, the structure model decides whether a page contains table-of-contents entries and whether the textbook body has begun. For contents pages it extracts pairs of the form \{title, level\}. Pages that are neither contents nor body are treated as cover or front matter and are discarded. Once the body is detected, a verification pass corrects invalid heading depths and removes entries that belong to prefaces or notation lists. If the model detects body text before extracting any valid contents entries, the whole module is retried, up to five attempts. M1 emits \texttt{toc.json}, \texttt{main\_text.md}, and the merged \texttt{images/} directory.

\paragraph{M2: hierarchy-conditioned segmentation.}
The recovered contents are traversed as a tree. Internal entries direct the traversal to their children, while leaf entries trigger a search for the corresponding body span. For each page and the current/next chapter titles, the model returns one of three states: \textsc{All-Current} (the page belongs to the current chapter), \textsc{All-Next} (it belongs to the next chapter), or \textsc{Boundary} (the transition occurs within the page). Boundary pages are checked a second time; a failed check is fed back to the assignment step. A successful transition advances the state machine and writes the completed chapter to \texttt{content/}.

\subsection{M3--M6: Exercise extraction and answer alignment}

\paragraph{M3: exercise separation.}
The structure model first tests whether the table of contents contains dedicated exercise chapters. When it does, those chapters are detached directly. When it does not, each content chapter is scanned for a trailing block of exercises, which is removed from the exposition. The same module constructs the multilevel exercise-title index and detects a separate answer chapter, if present. Its outputs are \texttt{content\_pure/}, \texttt{exercise/}, \texttt{exercise\_index.jsonl}, and optionally \texttt{answer.md}.

\paragraph{M4: exercise serialization.}
The parser reads each exercise section page by page and emits one JSON record per problem. A record contains \texttt{question}, \texttt{answer} when an answer is printed immediately after the question, and \texttt{number}. The final problem fragment on a page is carried into the next page before serialization so that a page break does not split a question into two records. Problem numbers are preserved for the matching stage.

\paragraph{M5: answer partitioning.}
The pipeline determines whether numbering is \emph{absolute}---one number identifies a unique problem in the whole book---or \emph{relative}, with numbers restarting across chapters. For absolute numbering, the answer chapter is serialized using the same page-wise procedure as M4. For relative numbering, the multilevel heading paths saved by M3 are used to partition the answer chapter before serialization.

\paragraph{M6: matching and filtering.}
Absolute-number records are joined directly on problem number. Relative-number records are matched within their recovered chapter, using the section content when necessary. Books without printed answers follow a third, unmatched branch. All branches then pass through a completeness check that removes questions whose statements depend on an unprovided theorem or on the conclusion of another exercise. The module produces a report with the number of exercises and answer coverage for each chapter.

\begin{figure}[t]
    \centering
    \resizebox{0.96\linewidth}{!}{%
    \begin{tikzpicture}[
        x=1cm,
        y=1cm,
        >=Latex,
        font=\small,
        rail/.style={draw=tbBlue!17, line width=5pt, line cap=round},
        flow/.style={->, draw=tbBlue, line width=0.9pt},
        bypass/.style={->, draw=tbInk!52, dashed, line width=0.75pt,
            rounded corners=5pt},
        stage/.style={circle, minimum size=9.5mm, inner sep=0pt,
            fill=tbBlue, text=white, font=\bfseries\small},
        decision/.style={diamond, aspect=1.35, minimum size=10.5mm,
            inner sep=0pt, fill=tbAmber, text=white, font=\bfseries\small},
        title/.style={font=\bfseries\small, text=tbInk, align=center},
        edgeword/.style={font=\scriptsize\bfseries, fill=white,
            inner sep=1.5pt},
        output/.style={draw=tbTeal, fill=tbTealLight, rounded corners=5pt,
            minimum width=2.9cm, minimum height=11mm, align=center,
            font=\bfseries\small, line width=0.85pt}
    ]
        \node[stage] (serialize) at (0.85,0) {1};
        \node[title, below=5mm of serialize] {Serialize problems};

        \node[decision] (answers) at (3.55,0) {2};
        \node[title, above=5mm of answers] {Answers available?};

        \node[decision] (scope) at (6.25,0) {3};
        \node[title, below=5mm of scope] {Numbering scope?};

        \node[stage] (align) at (8.95,0) {4};
        \node[title, above=5mm of align] {Align answers};

        \node[stage, fill=tbTeal] (check) at (11.65,0) {5};
        \node[title, below=5mm of check] {Standalone check};

        \node[output] (output) at (14.40,0) {Curated exercises};

        \draw[rail] (serialize.east) -- (answers.west);
        \draw[flow] (serialize.east) -- (answers.west);
        \draw[flow] (answers.east) --
            node[edgeword, text=tbBlue, above] {yes} (scope.west);

        \draw[->, draw=tbBlue, line width=0.85pt, rounded corners=4pt]
            (scope.east) -- ++(0.45,0) |- node[edgeword, text=tbBlue,
            near start, above] {global} (align.north);
        \draw[->, draw=tbBlue, line width=0.85pt, rounded corners=4pt]
            (scope.east) -- ++(0.45,0) |- node[edgeword, text=tbBlue,
            near start, below] {local} (align.south);

        \draw[rail] (align.east) -- (check.west);
        \draw[flow] (align.east) -- (check.west);
        \draw[->, draw=tbTeal, line width=0.95pt]
            (check.east) -- (output.west);

        \coordinate (no-route) at (3.55,-1.45);
        \draw[bypass] (answers.south) --
            node[edgeword, text=tbInk!62, left] {no} (no-route) -| (check.south);
    \end{tikzpicture}}
    \caption{Exercise evaluation track.  Problems are serialized, printed
    answers are aligned under the book's numbering scheme when available, and
    every candidate passes a standalone-completeness check.}
    \label{fig:appendix-exercise-pipeline}
\end{figure}
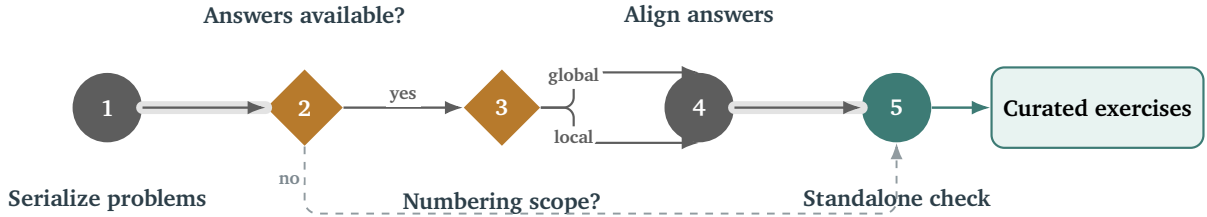

\subsection{Four-stage proposition pipeline}

Each pure-content chapter is first split at Markdown headings while preserving its preamble, and the resulting sections are processed concurrently. The four semantic stages are as follows.

\begin{enumerate}
    \item \textbf{Proposition extraction.}  For every section, the model emits
    structured proposition records with \texttt{type}, \texttt{number},
    \texttt{label}, \texttt{title}, \texttt{content\_tex}, and
    \texttt{proof\_tex} fields.
    \item \textbf{Reference resolution.}  A book-level proposition library is
    built from the extracted records.  For each proposition with a proof, the
    model identifies both explicit \texttt{\string\ref} citations and semantic
    references to earlier results, storing the matches in
    \texttt{resolved\_\allowbreak references}.
    \item \textbf{Proof-gap audit.}  Each proof is checked for unverified lemma
    assumptions, omitted coordinate or operator transformations, and unsupported
    jumps marked by phrases such as ``obviously'' or ``clearly.''  A candidate
    gap stores \texttt{gap\_\allowbreak statement},
    \texttt{missing\_\allowbreak logic},
    \texttt{sub\_\allowbreak proposition}, and a low/medium/high
    \texttt{difficulty} label.
    \item \textbf{Self-contained refinement.}  Each candidate sub-proposition is
    rewritten as an independent mathematical statement by declaring symbols,
    injecting required definitions, and preserving the source language.
\end{enumerate}

The chapter-level structured output contains the extracted propositions, resolved references, and candidate gaps. A separate Proof QA JSONL stream distinguishes original proposition records from candidate sub-propositions with an \texttt{is\_potential\_gap} flag. This schema describes the pipeline output; dataset-level acceptance, deduplication, and split policies are subsequent steps.

\begin{figure}[t]
    \centering
    \resizebox{0.96\linewidth}{!}{%
    \begin{tikzpicture}[
        x=1cm,
        y=1cm,
        >=Latex,
        font=\small,
        rail/.style={draw=tbBlue!17, line width=5pt, line cap=round},
        flow/.style={->, draw=tbBlue, line width=0.9pt},
        stage/.style={circle, minimum size=9.5mm, inner sep=0pt,
            fill=tbBlue, text=white, font=\bfseries\small},
        gap/.style={circle, minimum size=9.5mm, inner sep=0pt,
            fill=tbAmber, text=white, font=\bfseries\small},
        refine/.style={circle, minimum size=9.5mm, inner sep=0pt,
            fill=tbTeal, text=white, font=\bfseries\small},
        title/.style={font=\bfseries\small, text=tbInk, align=center},
        output/.style={draw=tbTeal, fill=tbTealLight, rounded corners=5pt,
            minimum width=2.65cm, minimum height=11mm, align=center,
            font=\bfseries\small, line width=0.85pt}
    ]
        \node[circle, fill=tbGrayLight, draw=tbInk!45, minimum size=11mm,
            inner sep=0pt] (content) at (0.75,0) {};
        \draw[draw=tbInk, line width=0.72pt]
            ($(content.center)+(-0.25,0.27)$) rectangle
            ($(content.center)+(0.25,-0.27)$);
        \draw[draw=tbInk!55, line width=0.52pt]
            ($(content.center)+(-0.16,0.10)$) -- ($(content.center)+(0.16,0.10)$);
        \draw[draw=tbInk!55, line width=0.52pt]
            ($(content.center)+(-0.16,-0.02)$) -- ($(content.center)+(0.16,-0.02)$);
        \draw[draw=tbInk!55, line width=0.52pt]
            ($(content.center)+(-0.16,-0.14)$) -- ($(content.center)+(0.08,-0.14)$);
        \node[title, below=5mm of content] {Section content};

        \node[stage] (extract) at (3.25,0) {1};
        \node[title, above=5mm of extract] {Extract propositions};

        \node[stage] (resolve) at (6.10,0) {2};
        \node[title, below=5mm of resolve] {Resolve references};

        \node[gap] (detect) at (8.95,0) {3};
        \node[title, above=5mm of detect] {Detect proof gaps};

        \node[refine] (refine) at (11.80,0) {4};
        \node[title, below=5mm of refine] {Refine obligations};

        \node[output] (output) at (14.55,0) {Proof supervision};

        \draw[rail] (content.east) -- (extract.west);
        \draw[rail] (extract.east) -- (resolve.west);
        \draw[rail] (resolve.east) -- (detect.west);
        \draw[flow] (content.east) -- (extract.west);
        \draw[flow] (extract.east) -- (resolve.west);
        \draw[flow] (resolve.east) -- (detect.west);
        \draw[->, draw=tbAmber, line width=0.95pt]
            (detect.east) -- (refine.west);
        \draw[->, draw=tbTeal, line width=0.95pt]
            (refine.east) -- (output.west);
    \end{tikzpicture}}
    \caption{Proposition-centered proof track.  Propositions are linked to
    their dependencies, local proof gaps are isolated, and selected obligations
    are refined into proof-supervision questions.}
    \label{fig:appendix-proposition-pipeline}
\end{figure}
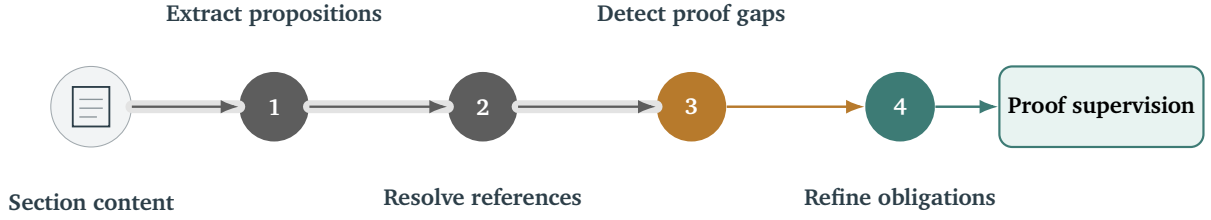

\subsection{Proposition extraction and model allocation}

The pure content chapters feed the proposition-centered training-data pipeline. The construction log compares three individual Gemini configurations and a hybrid allocation. The later cost-optimization note assigns second-stage reference semantics to the lower-cost non-thinking model and pairs it with the thinking model for the remaining reasoning-intensive operations. Because this allocation changed during pipeline development, it is versioned with each run. 
Table~\ref{tab:model-cost} reports the per-book costs recorded in the design document.

\begin{table}[H]
    \centering
    \small
    \caption{Model configurations considered for proposition extraction.  Cost
    estimates are those recorded during pipeline design.}
    \label{tab:model-cost}
    \renewcommand{\arraystretch}{1.08}
    \setlength{\aboverulesep}{0.6ex}
    \setlength{\belowrulesep}{0.8ex}
    \begin{tabular}{p{0.6\linewidth}p{0.15\linewidth}}
        \toprule
        \textbf{Configuration} & 
        \textbf{Cost / book} \\
        \midrule
        \textit{Config 1:} \texttt{gemini-3.1-pro-preview-thinking}  & \$40--50 \\
        \textit{Config 2:} \texttt{gemini-3.1-flash-lite-preview}  & approx.\ \$5 \\
        \textit{Config 3:} \texttt{gemini-3-flash-preview-thinking}  & approx.\ \$10 \\
        \textit{Config 2 + Config 3} hybrid & approx.\ \$7 \\
        \bottomrule
    \end{tabular}
\end{table}

\subsection{Model-Quality Evaluation and Model Selection}
\label{app:model_quality}

\paragraph{Evaluation Dimensions.}
We evaluate model quality along six dimensions: 
\begin{itemize}
    \item \textit{Completeness} measures the completeness of extracted information;
    \item \textit{Accuracy} assesses factual correctness;
    \item \textit{LaTeX Formatting} evaluates compliance with LaTeX formatting requirements;
    \item \textit{Self-Containment} measures whether each extracted item is independently understandable;
    \item \textit{Atomicity} assesses whether each item represents a single, coherent unit
    \item \textit{C/P Ratio} measures cost-effectiveness in terms of the cost of generating a single QA pair, with a higher score indicating a lower cost per pair.
\end{itemize}

\paragraph{Model Configurations.}
We compare four configurations: 
\begin{itemize}
    \item \textit{Flash Thinking} refers to \texttt{gemini-3-flash-preview-thinking}
\item \textit{Flash Lite} refers to \texttt{gemini-3.1-flash-lite-preview}
\item \textit{Pro Thinking} refers to \texttt{gemini-3.1-pro-preview-thinking}
\item \textit{Combined} denotes a pipeline combining Flash Thinking for primary extraction and auditing with Flash Lite for reference-semantic parsing.
\end{itemize}

\paragraph{Results and Analysis.}
Figure~\ref{fig:data-reserve-quality} shows that Pro Thinking achieves strong performance in Completeness, Accuracy, LaTeX Formatting, Self-Containment, and Atomicity, but performs less favorably in C/P Ratio, indicating a higher cost per generated QA pair. Flash Lite offers a more cost-effective alternative, although its lower scores in Completeness and Self-Containment suggest potential limitations in information coverage and the independent interpretability of extracted items. Flash Thinking provides consistently strong performance across the five quality-oriented dimensions, while maintaining a reasonable cost per QA pair. Notably, the Combined pipeline preserves competitive extraction quality while improving cost-effectiveness through the use of Flash Lite for reference-semantic parsing. Overall, these results suggest that the Combined pipeline offers a favorable balance between extraction quality and the cost of generating individual QA pairs.

\section{Extended Characterization Analysis}
\label{sec:characterization-appendix}

This appendix provides the detailed corpus characterization summarized in
Section~\ref{sec:characterization}.  We first describe subject and language
coverage and task composition, then examine quality behavior, provenance,
reasoning scope, structural complexity, and confidence.

\subsection{Task and Answer-Format Composition}
\label{sec:appx-task}

\begin{figure}[H]
    \centering
    \includegraphics[width=0.8\linewidth]{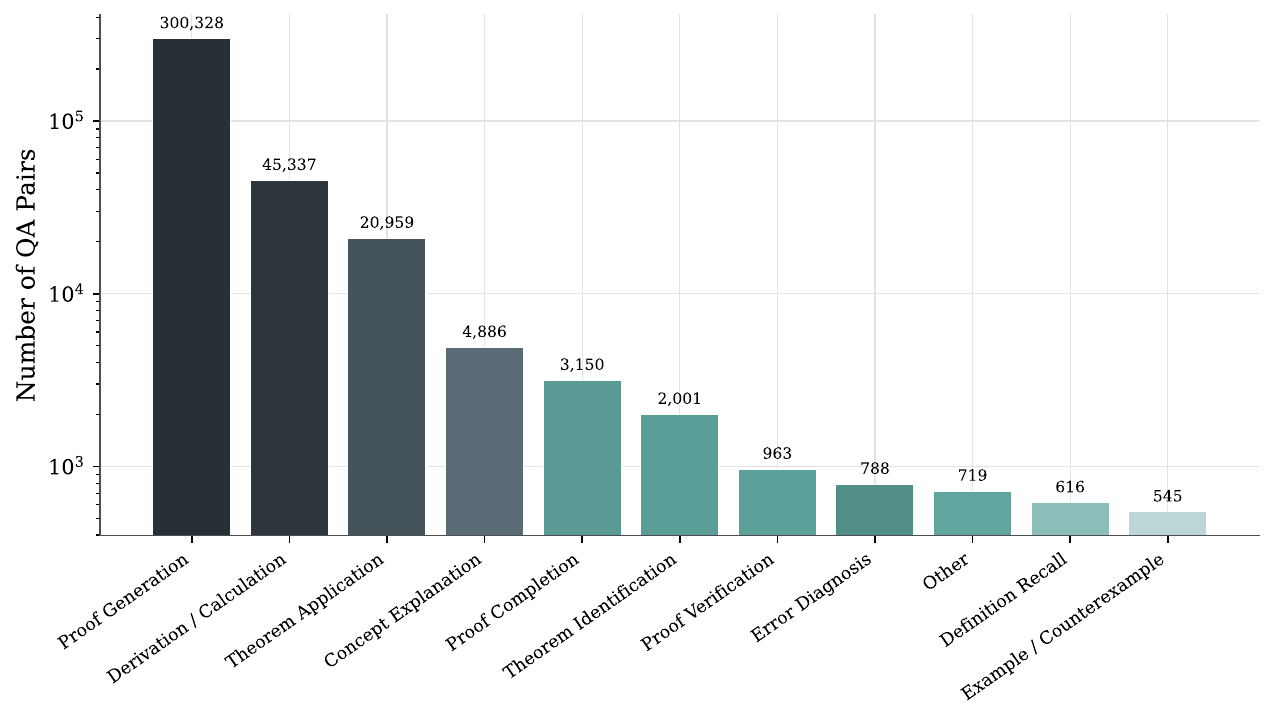}
    \par\smallskip
    \includegraphics[width=0.76\linewidth]{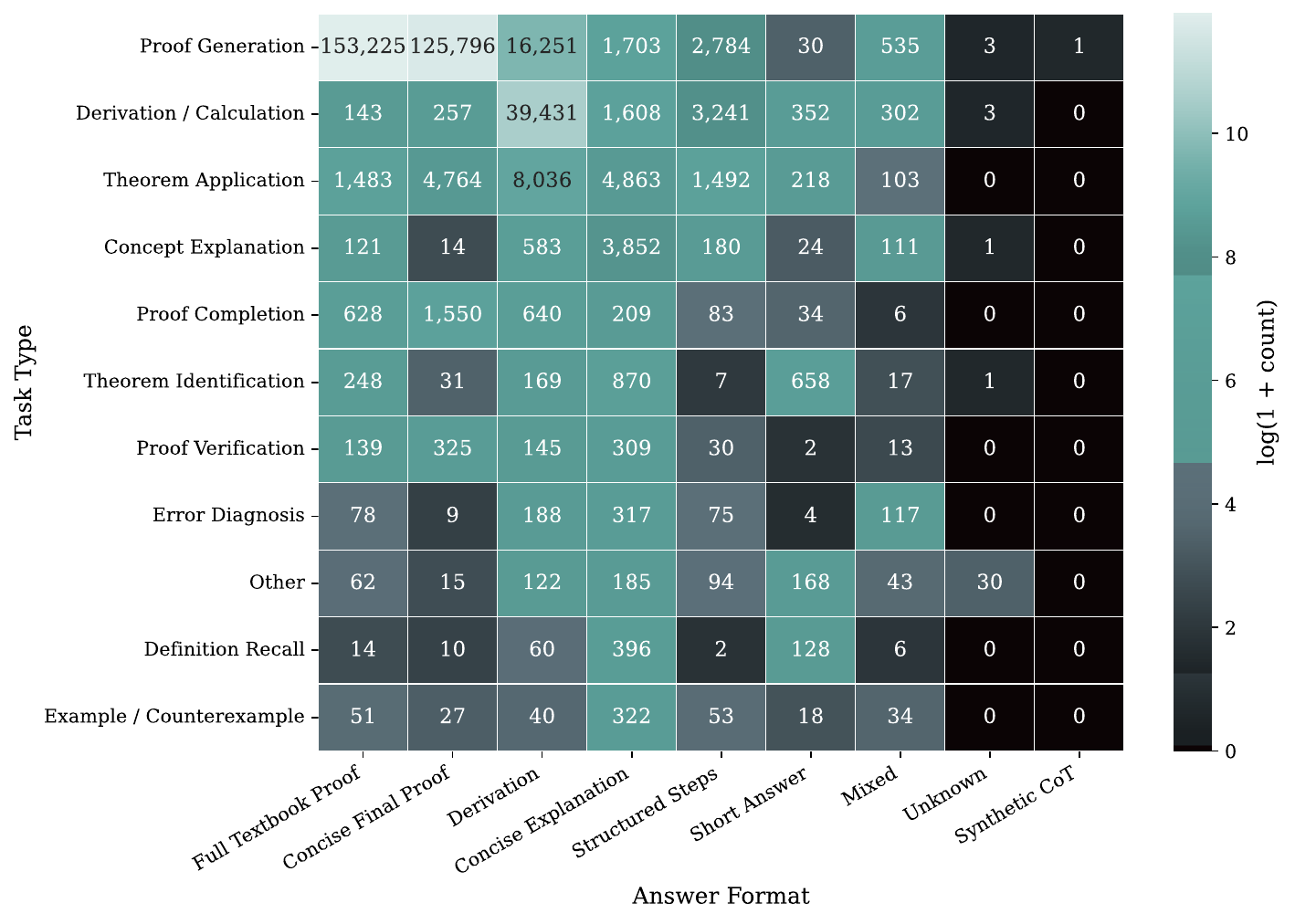}
    \caption{(a) Distribution of eleven fine grained task types on a
    logarithmic count axis. (b) Joint distribution of task type and answer
    format on a logarithmic color scale.}
    \label{fig:appx-task}
\end{figure}

Figure~\ref{fig:appx-task}(a) shows that proof generation accounts for 79.0\%
of records, an order of magnitude above every other task type.
Figure~\ref{fig:appx-task}(b) crosses task type with answer format. Derivation
and calculation binds almost deterministically to the derivation format at
87.0\%, while proof generation splits between full textbook proof (51.0\%)
and concise final proof (41.9\%).

\subsection{Subject and Language Coverage}
\label{sec:appx-subject}

\begin{figure}[H]
    \centering
    \includegraphics[width=\linewidth]{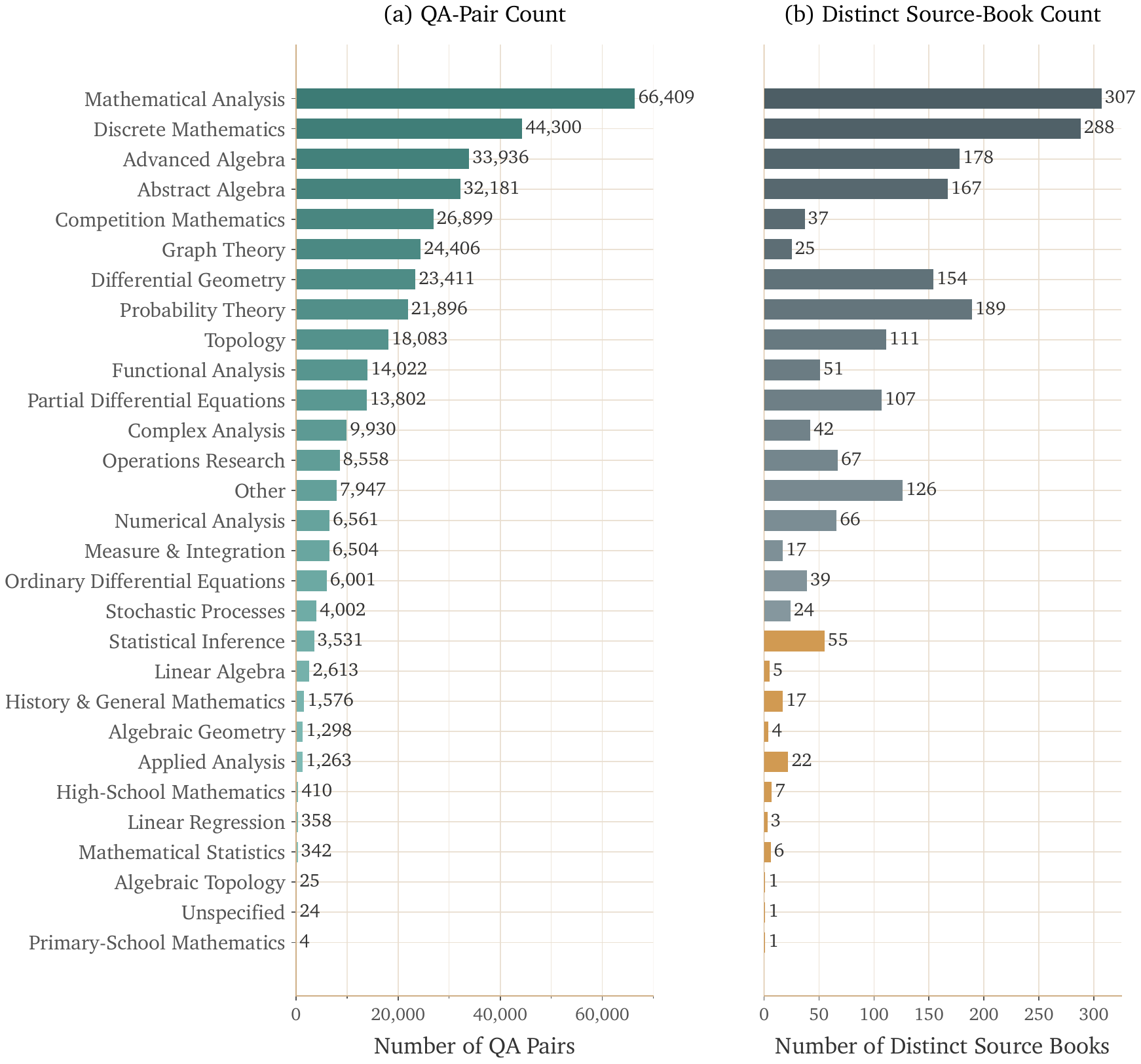}
    \caption{(a) QA volume and (b) distinct source book count across all
    subjects, sharing a common subject ordering.}
    \label{fig:appx-subject-lang}
\end{figure}

Figure~\ref{fig:appx-subject-lang}(a)-(b) reports QA volume and distinct
source book count for the fifteen most represented subjects. The ten leading
subjects account for 84.4\% of records. Mathematical analysis draws 66{,}409
records from 307 books (216 per book), while graph theory draws 24{,}406
records from only 25 books (976 per book), reflecting broad coverage for
analysis-heavy subjects against concentration on specialized books elsewhere.
Source language composition varies by educational level. All 337 high school
records have mixed-language sources. Undergraduate records comprise
93{,}537 Chinese, 71{,}702 English, and 23{,}500 mixed-language examples;
competition records comprise 24{,}538 Chinese (92.4\%) and 2{,}024
mixed-language examples; and graduate records comprise 109{,}648 English
(66.6\%), 35{,}967 Chinese, and 19{,}039 mixed-language examples.

\thispagestyle{fancy}
\FloatBarrier
\clearpage

\subsection{Quality Behavior at Subject and Dimension Granularity}
\label{sec:appx-quality}

\begin{figure}[H]
    \centering
    \includegraphics[width=0.66\linewidth]{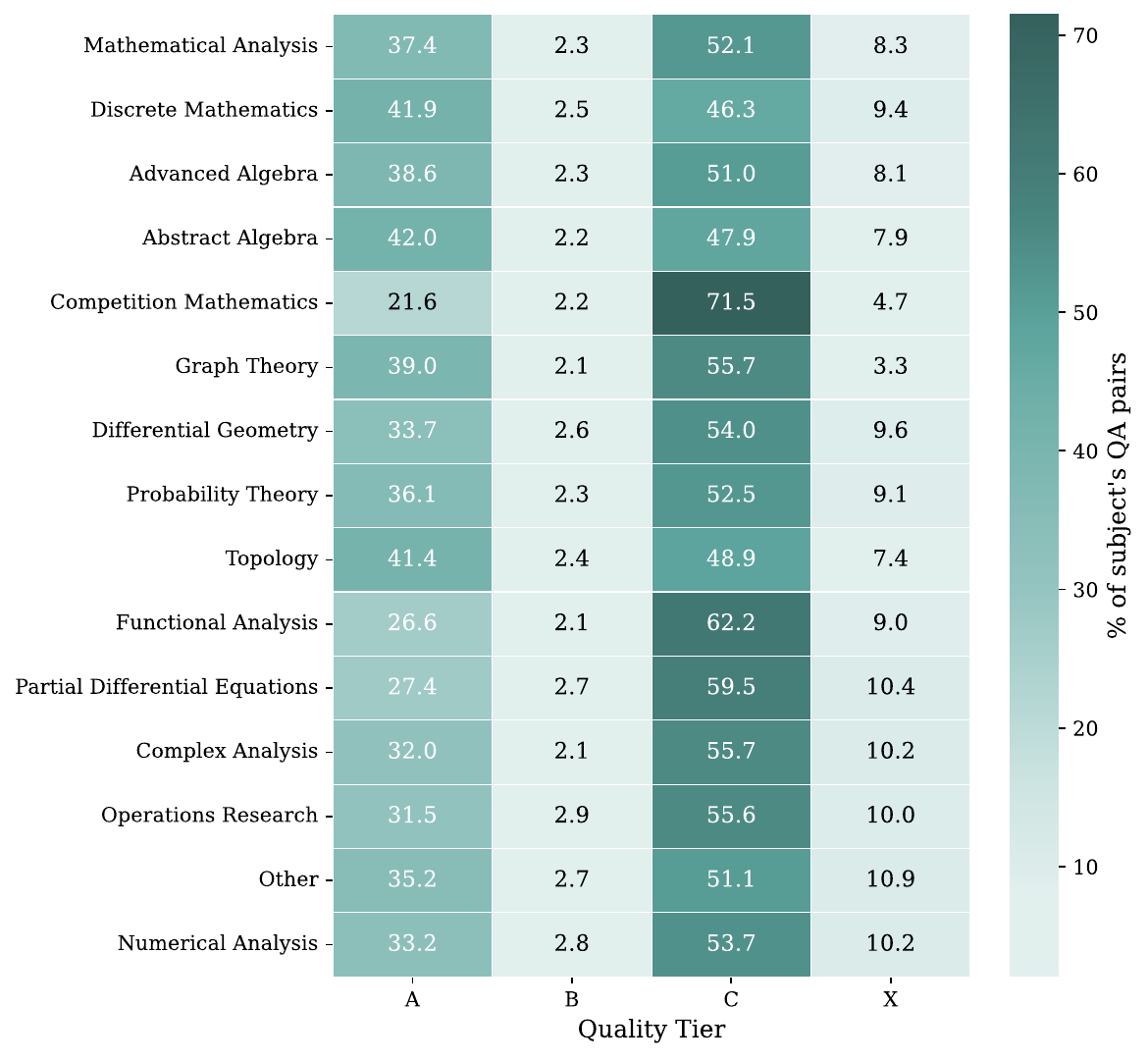}
    \par\medskip
    \includegraphics[width=0.78\linewidth]{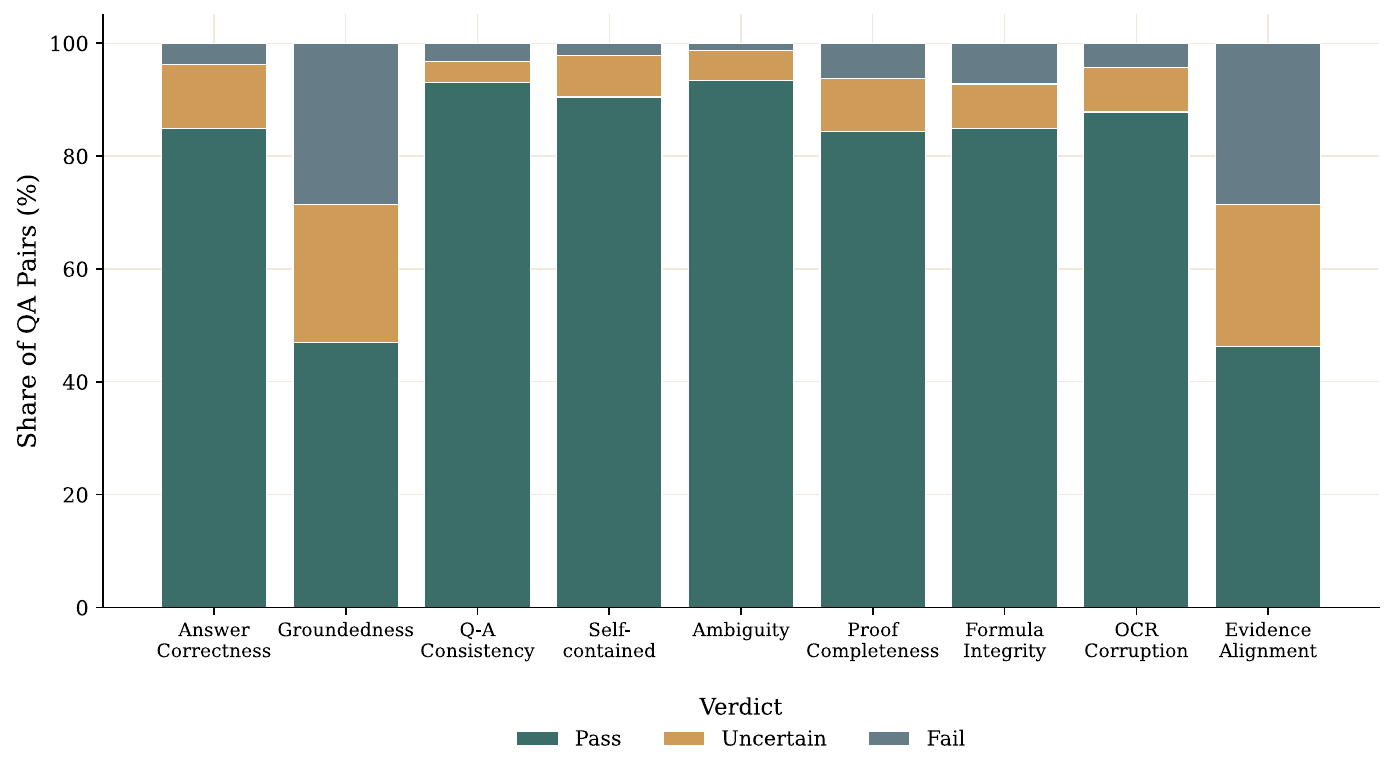}
    \caption{Quality outcomes decomposed by (a) subject, normalized by row,
    and (b) pass, uncertain, and fail rates across nine audit dimensions.}
    \label{fig:appx-quality}
\end{figure}

Figure~\ref{fig:appx-quality}(a) presents subject-specific quality-tier
profiles. Competition mathematics has tier A and tier C shares of 21.6\%
and 71.5\%, respectively, while abstract algebra and topology each have tier A
shares above 39\%. Figure~\ref{fig:appx-quality}(b) reports results across nine
automated audit dimensions. Ambiguity and question--answer consistency pass
above 90\% of records. Groundedness and evidence alignment assess correspondence
with the source material; both pass below 50\% of records and have a fail rate
of 28.6\%. This dimension-level breakdown provides a detailed profile of
question--answer quality and source-evidence traceability to guide targeted
review.

\begin{figure}[H]
    \centering
    \includegraphics[width=0.90\linewidth]{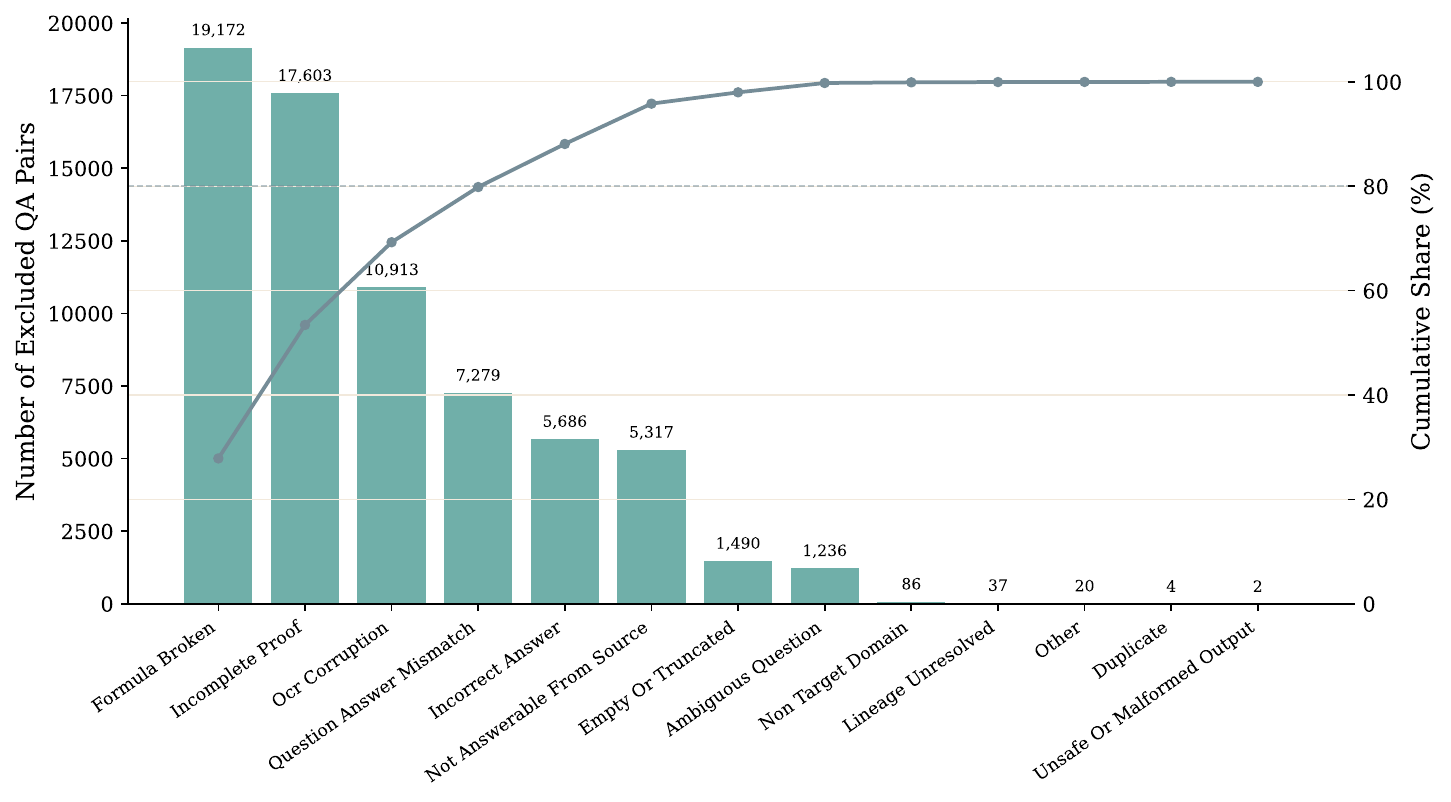}
    \par\medskip
    \includegraphics[width=0.90\linewidth]{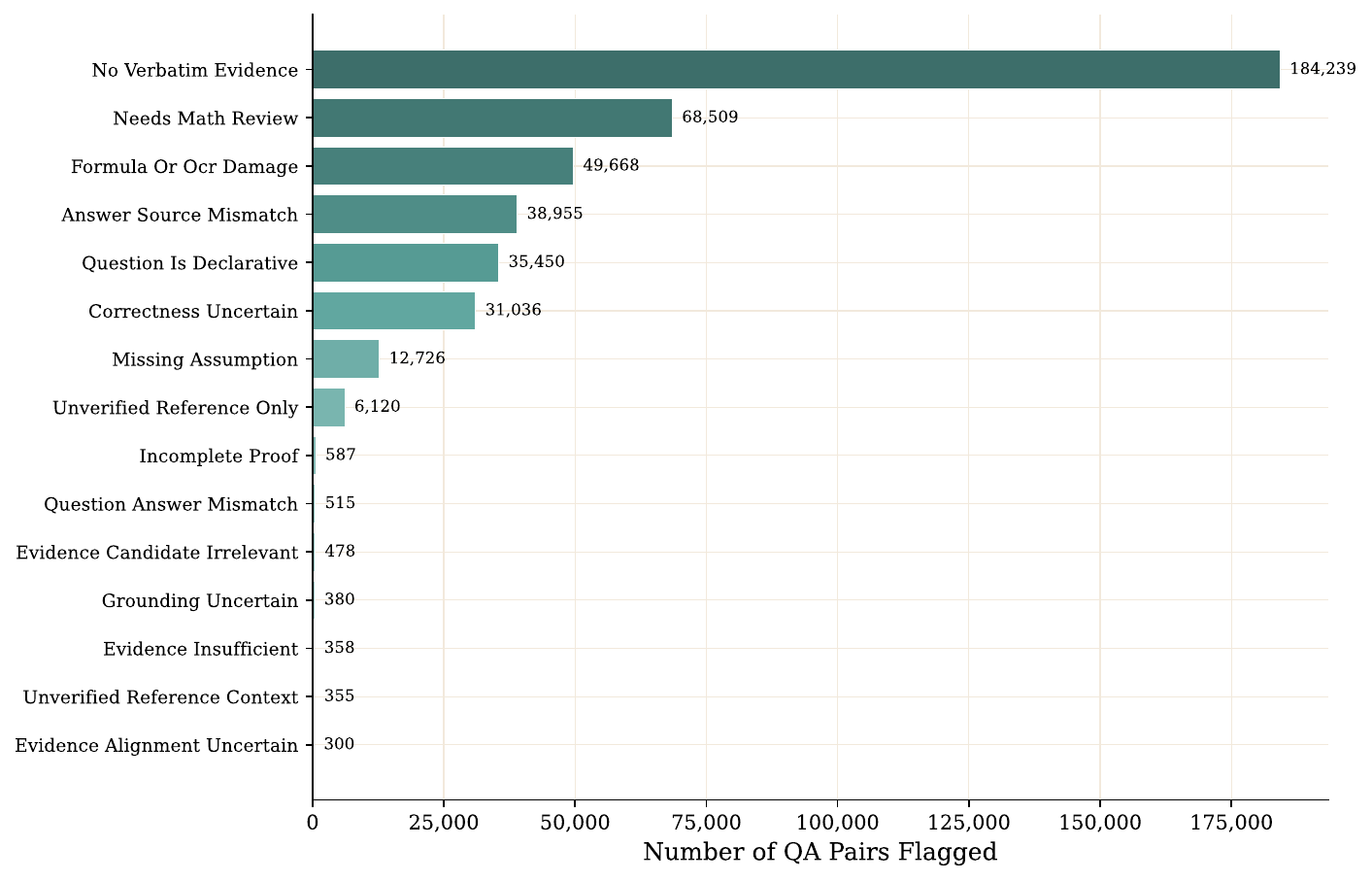}
    \caption{(a) Pareto distribution of record exclusion reasons, with
    cumulative share on the secondary axis. (b) Fifteen leading human review
    flags, with multiple flags permitted per record.}
    \label{fig:appx-reasons}
\end{figure}

Figure~\ref{fig:appx-reasons}(a) reports the reasons records are excluded outright. Formula broken and incomplete proof are the two leading causes, together accounting for 36{,}775 excluded records, with the top four reasons covering 80\% of all exclusions. Figure~\ref{fig:appx-reasons}(b) corroborates the dimension level asymmetry in Figure~\ref{fig:appx-quality}(b) from an independent signal drawn from human review flags. The flag no verbatim evidence covers 48.4\% of the corpus, 2.7 times the frequency of the second most common flag, needs math review.

\FloatBarrier

\subsection{Corpus Redundancy and Generation Provenance}
\label{sec:appx-provenance}

\begin{figure}[!htbp]
    \centering
    \begin{minipage}[c]{0.44\linewidth}
        \centering
        \includegraphics[width=\linewidth]{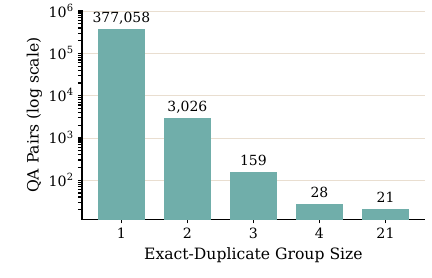}
    \end{minipage}\hfill
    \begin{minipage}[c]{0.52\linewidth}
        \centering
        \includegraphics[width=\linewidth]{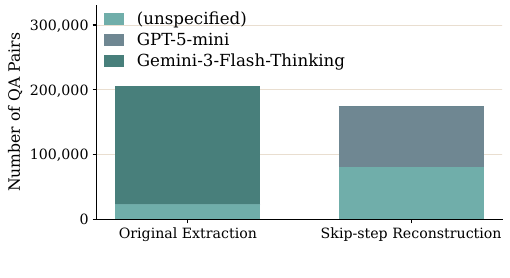}
    \end{minipage}
    \caption{(a) Exact duplicate group sizes on a logarithmic count axis.
    (b) Generator model assignment by generation stream.}
    \label{fig:appx-redundancy}
\end{figure}

Figure~\ref{fig:appx-redundancy}(a) reports exact duplicate group sizes. Singleton records account for 99.15\% of the corpus and duplicate groups account for only 0.85\%, with one outlier group of size twenty one isolated from the remaining groups, all of size four or smaller. Figure~\ref{fig:appx-redundancy}(b) traces the generation composition introduced in Section~\ref{sec:char-scale} to individual generator models. The Proof QA and Gap QA streams draw on disjoint model assignments, with \texttt{gemini-3-flash-preview-thinking} responsible for 88.9\% of Proof QA records and \texttt{gpt-5-mini} responsible for 53.7\% of Gap QA records.

\begin{figure}[H]
    \centering
    \includegraphics[width=\linewidth]{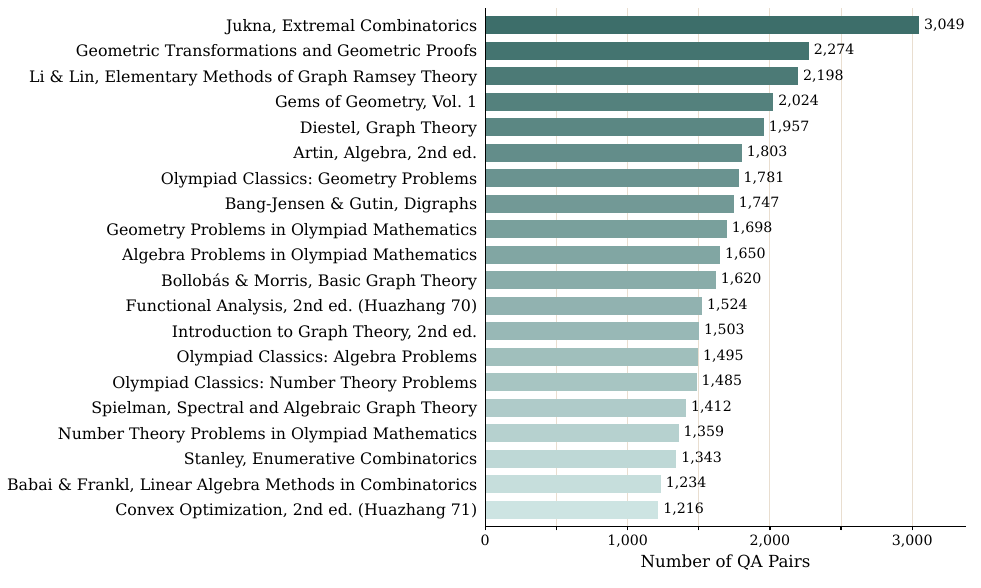}
    \caption{Twenty highest yield source books out of 2{,}118 distinct books in the corpus.}
    \label{fig:appx-top-books}
\end{figure}

Figure~\ref{fig:appx-top-books} traces the subject-level concentration in
Section~\ref{sec:appx-subject} to individual sources. The single highest yield book, Jukna's Extremal Combinatorics, contributes 3{,}049 records, 0.80\% of the corpus, and the top twenty books are dominated by English language graph theory and combinatorics textbooks together with Chinese competition mathematics series.

\FloatBarrier

\subsection{Reasoning Scope and Subject Composition Across Level}
\label{sec:appx-level}

\begin{figure}[H]
    \centering
    \includegraphics[width=\linewidth]{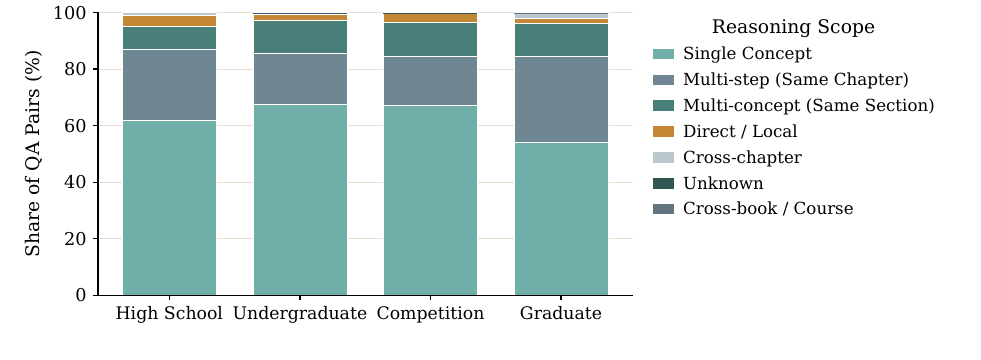}
    \par\medskip
    \includegraphics[width=\linewidth]{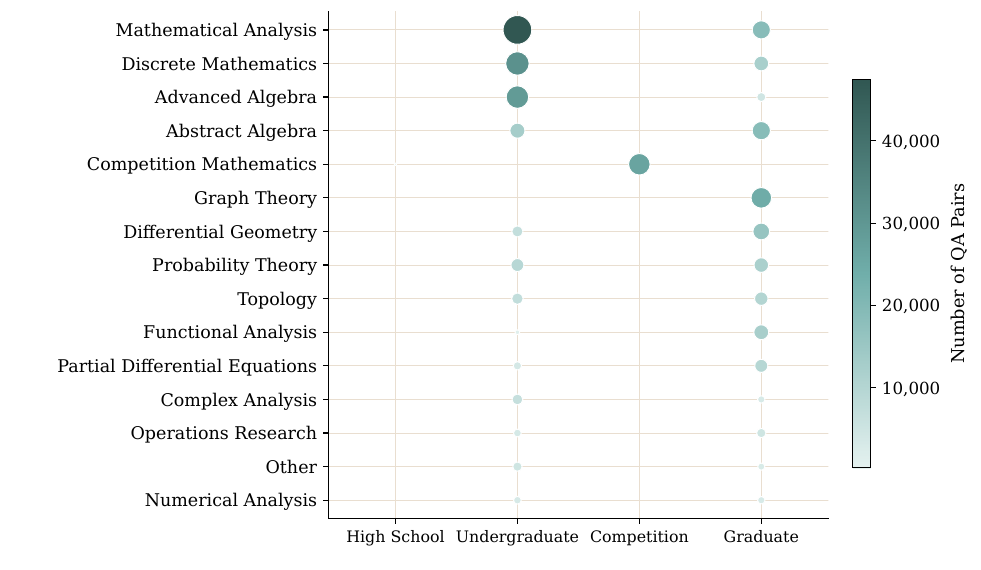}
    \caption{(a) Percentage composition of reasoning scope within each
    educational level. (b) Joint distribution of the fifteen most represented
    subjects and educational level.}
    \label{fig:appx-level}
\end{figure}

Figure~\ref{fig:appx-level}(a) examines how reasoning scope, the breadth of chapters and concepts a record draws on, varies with educational level. The combined share of multi step same chapter and cross chapter reasoning reaches 32.5\% at the graduate level, compared with 18.7\% for undergraduate and 17.7\% for competition. Competition and undergraduate report similar single concept shares despite differing in solving difficulty, indicating that reasoning scope captures curriculum breadth rather than solving skill. Figure~\ref{fig:appx-level}(b) places this finding in the context of subject coverage from Section~\ref{sec:appx-subject}. Mathematical analysis is the only leading subject with substantial coverage at both undergraduate and graduate levels, while most other leading subjects concentrate on a single level, explaining why level and subject interact strongly in downstream evaluation splits.

\FloatBarrier

\subsection{Structural Complexity and Confidence Dynamics}
\label{sec:appx-structure-confidence}

\begin{figure}[!htbp]
    \centering
    \includegraphics[width=\linewidth]{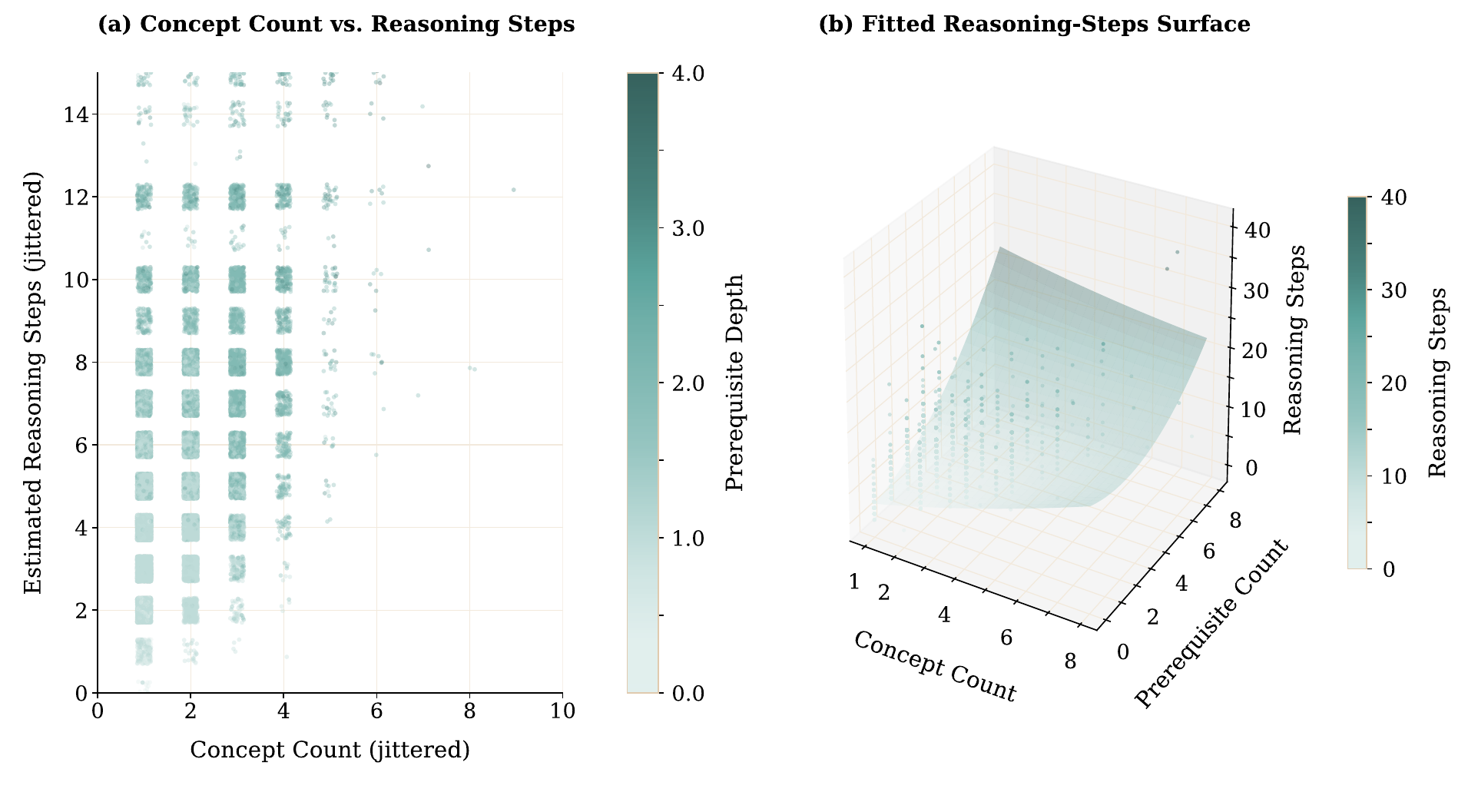}
    \par\vspace{10pt}
    \includegraphics[width=\linewidth]{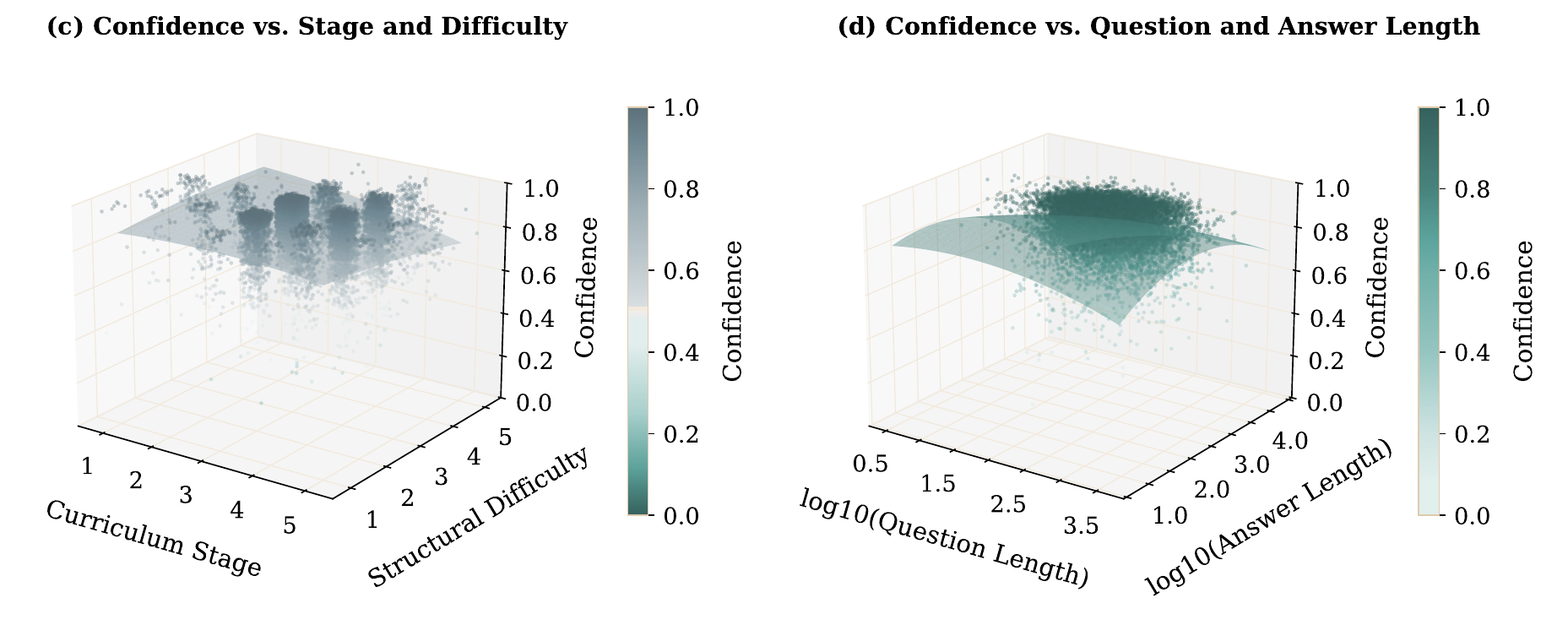}
    \caption{(a) Joint distribution of concept count and estimated reasoning
    steps, colored by prerequisite depth. (b) Fitted surface of estimated
    reasoning steps over concept count and prerequisite count. (c) Fitted
    surface of model confidence over curriculum stage and structural
    difficulty. (d) Fitted surface of model confidence over log-scaled
    question length and answer length.}
    \label{fig:appx-structure-confidence}
\end{figure}

Each record carries structural metadata (curriculum stage, structural
difficulty, concept count, prerequisite depth) and a model-reported confidence
score. Figure~\ref{fig:appx-structure-confidence}(a)-(b) shows that
prerequisite count correlates with estimated reasoning steps at 0.633,
stronger than concept count at 0.522; the fitted surface also rises more
steeply along the prerequisite-count axis, making prerequisite depth the
dominant driver of reasoning-chain length.
Figure~\ref{fig:appx-structure-confidence}(c) crosses confidence with
curriculum stage and difficulty. Confidence falls with difficulty only at
later stages, from 0.876 to 0.807 at stage four and from 0.873 to 0.788 at
stage five, while stages one through three remain flat.
Figure~\ref{fig:appx-structure-confidence}(d) shows that answer length
correlates with confidence at $-0.231$, compared with $-0.060$ for question
length, suggesting that longer derivations drive lower confidence.

\FloatBarrier

\subsection{Structural Difficulty and Record Length}
\label{sec:appx-structure}

\begin{figure}[H]
    \centering
    \begin{minipage}[c]{0.48\linewidth}
        \centering
        \includegraphics[width=\linewidth]{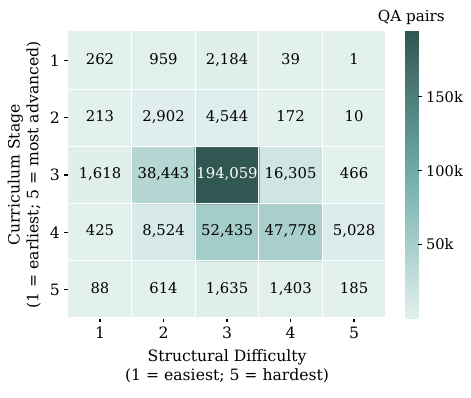}
    \end{minipage}\hfill
    \begin{minipage}[c]{0.48\linewidth}
        \centering
        \includegraphics[width=\linewidth]{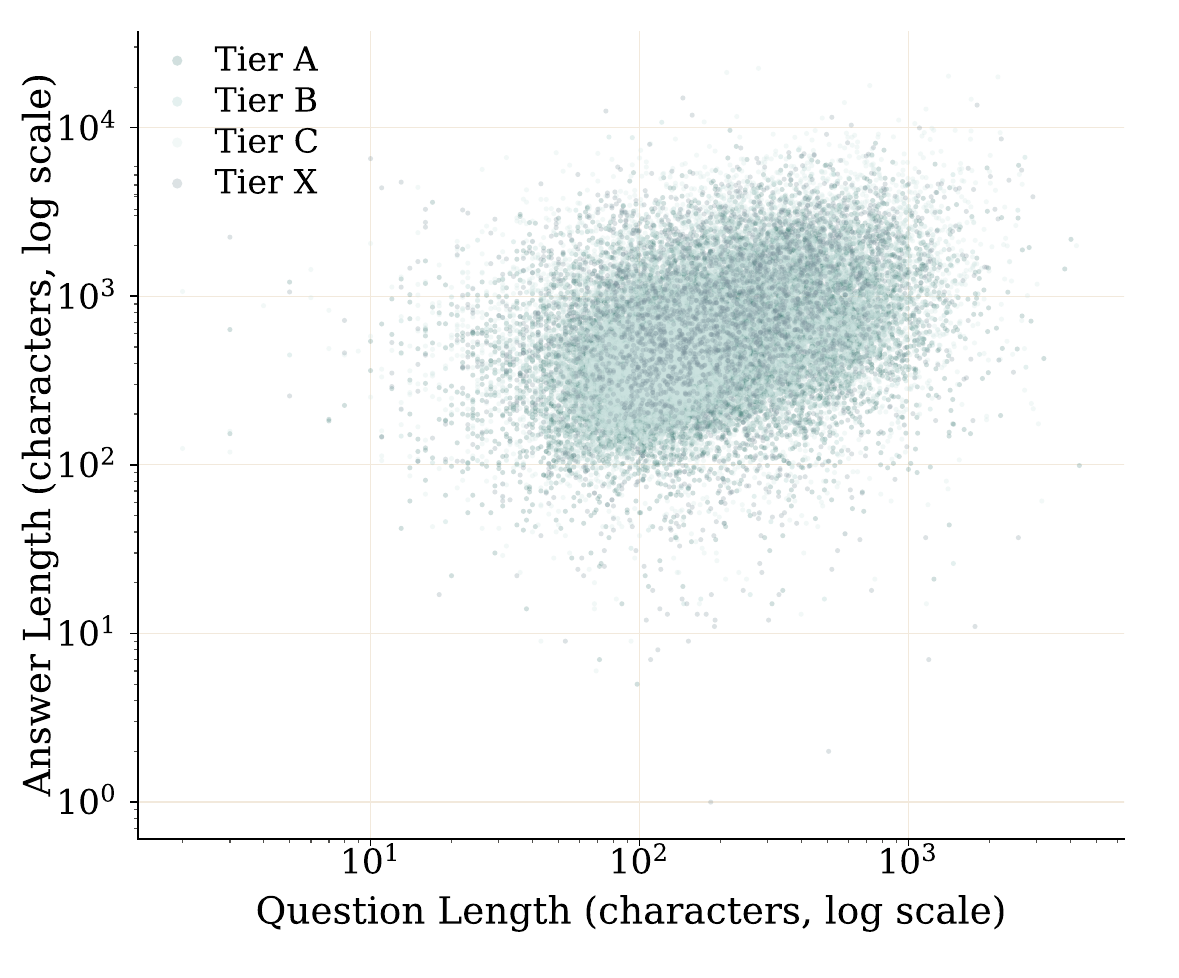}
    \end{minipage}
    \caption{(a) Joint distribution of curriculum stage and structural
    difficulty. (b) Question and answer length by quality tier, on logarithmic
    axes.}
    \label{fig:appx-structure}
\end{figure}

Figure~\ref{fig:appx-structure}(a) reports the underlying joint distribution
of curriculum stage and structural difficulty. A single cell at stage three, difficulty three holds 194{,}059 records, 51.0\% of the corpus, and stage three overall accounts for 65.9\% of records, indicating that structural difficulty tracks curriculum progression closely rather than varying freely within a stage. Figure~\ref{fig:appx-structure}(b) reports the joint distribution of question and answer length by quality tier. The two lengths are positively associated but retain substantial spread within every tier, so tier assignment cannot be inferred from length alone.

\FloatBarrier

\subsection{Confidence Distributions by Tier and Curriculum Stage}
\label{sec:appx-confidence}

\begin{figure}[H]
    \centering
    \begin{minipage}[c]{0.48\linewidth}
        \centering
        \includegraphics[width=\linewidth]{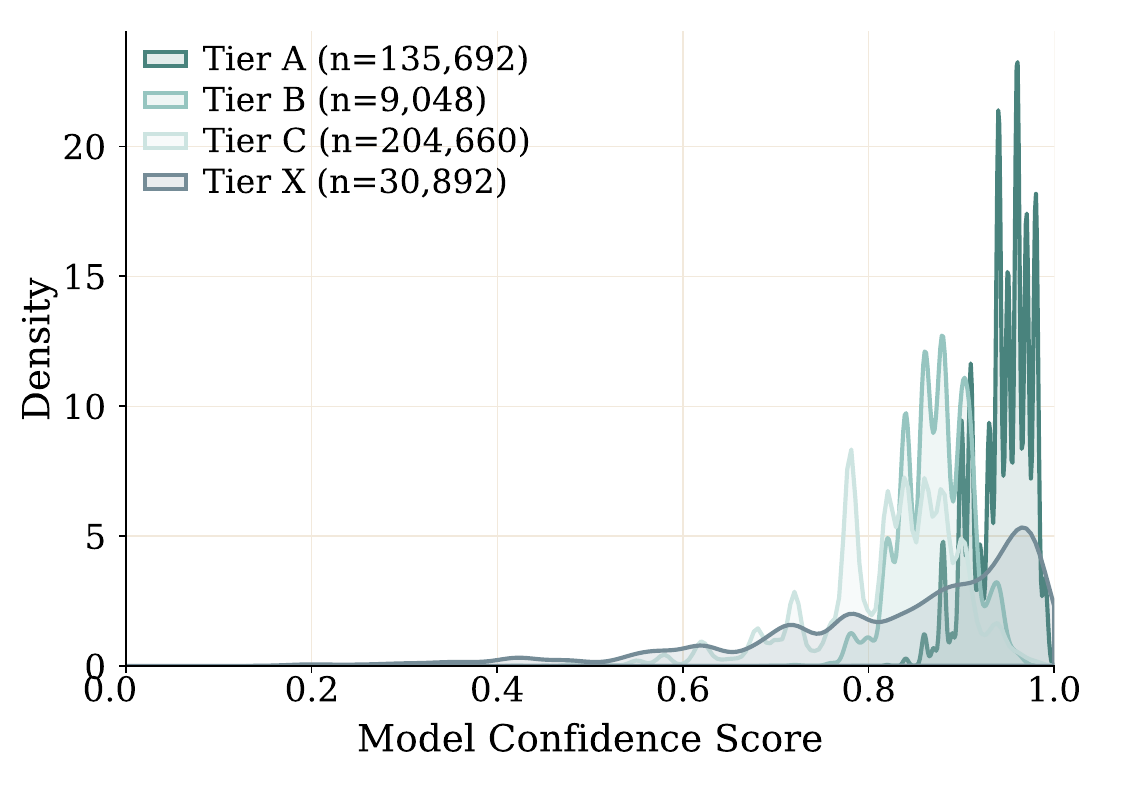}
    \end{minipage}\hfill
    \begin{minipage}[c]{0.48\linewidth}
        \centering
        \includegraphics[width=\linewidth]{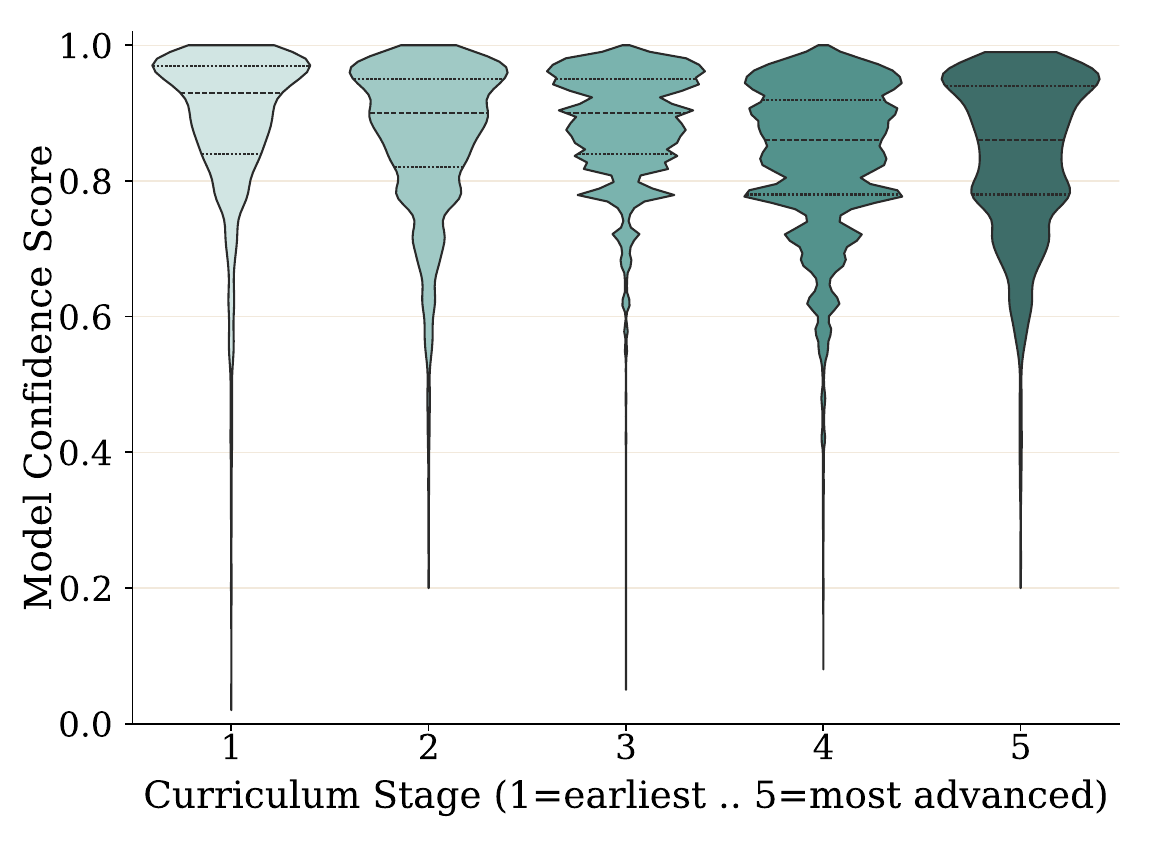}
    \end{minipage}
    \caption{(a) Kernel density estimate of model-reported confidence by
    quality tier. (b) Distribution of model confidence across the five
    curriculum stages.}
    \label{fig:appx-confidence}
\end{figure}

Figure~\ref{fig:appx-confidence}(a) reports the marginal confidence
distribution by quality tier. Mean confidence follows the order tier A at 0.944, tier B at 0.873, and tier C at 0.821, with tier A showing the tightest spread at a standard deviation of 0.030, indicating that confidence separates tier A cleanly from the lower tiers. Figure~\ref{fig:appx-confidence}(b) reports the marginal distribution by curriculum stage, which follows a non monotonic pattern of 0.892, 0.865, 0.882, 0.835, and 0.840 across stages one through five, with stage three holding both the largest sample share at 66.0\% of the corpus and the tightest spread.

\FloatBarrier

\section{Synthesized QA Samples}
\label{app:qa-samples}

To make the pipeline output concrete, this appendix pairs each synthesized QA with the exact source paragraph it was derived from. We draw all five items from a single textbook---Diestel's \emph{Graph Theory}, one of the graph-theory books in the corpus---so that the reader can trace how the same book supplies items at every level of the L1--L5 ladder used in Section~\ref{sec:characterization}. Each block below reproduces the source excerpt (grey) and the QA record (green) verbatim from the intermediate representation and the release table; only whitespace, LaTeX escaping, and page-break markers were adjusted for typesetting.

\newtcolorbox{srcbox}[1]{%
    colback=gray!5, colframe=gray!40,
    boxrule=0.5pt, arc=1pt, left=6pt, right=6pt, top=5pt, bottom=5pt,
    breakable, fonttitle=\small\bfseries, title={#1}
}
\newtcolorbox{qabox}[1]{%
    colback=tbTealLight, colframe=tbTeal!60,
    colbacktitle=tbTeal, coltitle=white,
    boxrule=0.5pt, arc=1pt, left=6pt, right=6pt, top=5pt, bottom=5pt,
    breakable, fonttitle=\small\bfseries, title={#1}
}

\subsection*{L1 -- Direct fact recall (Section 12.3, Tree-decompositions)}

\begin{srcbox}{Source excerpt}
\small
Let $G$ be a graph, $T$ a tree, and let $\mathcal{V} = (V_t)_{t\in T}$ be a family of \emph{bags} $V_t \subseteq V(G)$ indexed by the nodes $t$ of $T$. The pair $(T,\mathcal{V})$ is called a \emph{tree-decomposition} of $G$ if it satisfies the following three conditions:
\begin{itemize}\itemsep0pt
    \item[(T1)] $V(G) = \bigcup_{t\in T} V_t$;
    \item[(T2)] for every edge $e \in G$ there exists a $t\in T$ such that both ends of $e$ lie in $V_t$;
    \item[(T3)] $V_{t_1} \cap V_{t_3} \subseteq V_{t_2}$ whenever $t_1, t_2, t_3 \in T$ satisfy $t_2 \in t_1 T t_3$.
\end{itemize}
\end{srcbox}

\begin{qabox}{Synthesized QA}
\small
\textbf{Question.}
Let $G$ be a graph and let $(T,\mathcal{V})$ be a tree decomposition of $G$, where $\mathcal{V} = \{V_t : t \in V(T)\}$ is a family of bags such that $\bigcup_{t\in T} V_t = V(G)$. Given a set of vertices $W \subseteq V(G)$, prove that for any $w\in W$ there exists a node $t' \in V(T)$ with $w \in V_{t'}$.

\smallskip
\textbf{Answer.}
By definition of a tree decomposition (condition (T1)), $V(G) = \bigcup_{t\in V(T)} V_t$, i.e.\ the bags cover all vertices of $G$. Since $W \subseteq V(G)$, every $w \in W$ lies in $V(G) = \bigcup_{t\in V(T)} V_t$, so there exists a node $t' \in V(T)$ with $w \in V_{t'}$. \hfill$\square$
\end{qabox}

\subsection*{L2 -- Concept explanation (Section 5.4, List colouring)}

\begin{srcbox}{Source excerpt}
\small
In principle, showing that a given graph is $k$-choosable is more difficult than proving it to be $k$-colourable: the latter is just the special case of the former where all lists are equal to $\{1,\ldots,k\}$. Thus,
\[ \mathrm{ch}(G) \geq \chi(G) \quad \text{and} \quad \mathrm{ch}'(G) \geq \chi'(G) \]
for all graphs $G$.
\end{srcbox}

\begin{qabox}{Synthesized QA}
\small
\textbf{Question.}
Explain, in the setting of list colouring, why $\mathrm{ch}(G) \geq \chi(G)$ and $\mathrm{ch}'(G) \geq \chi'(G)$ for every graph $G$.

\smallskip
\textbf{Answer.}
In principle, showing that a given graph is $k$-choosable is more difficult than proving it to be $k$-colourable: the latter is just the special case of the former where all lists are equal to $\{1,\ldots,k\}$. Hence any list of size $k$ can be specialized to this uniform list, and the two inequalities follow.
\end{qabox}

\subsection*{L3 -- Single-concept short proof (Section 1.2, Proposition 1.2.1)}

\begin{srcbox}{Source excerpt}
\small
Summing $d(v)$ over all $v \in V$ counts every edge twice, so
\[ |E| = \tfrac{1}{2}\sum_{v\in V} d(v). \]

\textbf{Proposition 1.2.1.} \emph{The number of vertices of odd degree in a graph is always even.}

\emph{Proof.} As $|E| = \tfrac{1}{2}\sum_{v\in V} d(v)$ is an integer, $\sum_{v\in V} d(v)$ is even. \hfill$\square$
\end{srcbox}

\begin{qabox}{Synthesized QA}
\small
\textbf{Question.}
Prove that the number of vertices of odd degree in a graph is always even.

\smallskip
\textbf{Answer.}
Since $|E| = \tfrac{1}{2}\sum_{v\in V} d(v)$ is an integer, $\sum_{v\in V} d(v)$ is even. Reducing this sum modulo $2$, each even-degree vertex contributes $0$ and each odd-degree vertex contributes $1$, so $|V_{\mathrm{odd}}| \equiv 0 \pmod 2$; the set of odd-degree vertices has even cardinality. \hfill$\square$
\end{qabox}

\subsection*{L4 -- Chapter-local multi-step proof (Section 1.9, Lemma 1.9.3)}

\begin{srcbox}{Source excerpt}
\small
\textbf{Proposition 1.9.2.} Together with $\emptyset$, the cuts in $G$ form a subspace $\mathcal{B}(G)$ of $\mathcal{E}(G)$. This space is generated by atomic cuts. \emph{(A minimal non-empty cut is a bond.)}

\smallskip
\textbf{Lemma 1.9.3.} \emph{Every cut is a disjoint union of bonds.}

\emph{Proof.} We apply induction on the size of the cut $F$ considered. For $F = \emptyset$ the assertion is trivial (with the empty union). If $F \neq \emptyset$ is not itself a bond, it properly contains some other non-empty cut $F'$. By Proposition~1.9.2, $F \setminus F' = F + F'$ is also a smaller non-empty cut. By the induction hypothesis, both $F'$ and $F \setminus F'$ are disjoint unions of bonds, and hence so is $F$. \hfill$\square$
\end{srcbox}

\begin{qabox}{Synthesized QA}
\small
\textbf{Question.}
Prove that every cut in a graph is a disjoint union of bonds.

\smallskip
\textbf{Answer.}
We apply induction on the size of the cut $F$. For $F = \emptyset$ the assertion is trivial (with the empty union). If $F \neq \emptyset$ is not itself a bond, it properly contains some other non-empty cut $F'$. By Proposition~1.9.2, $F \setminus F' = F + F'$ is also a smaller non-empty cut. By the induction hypothesis, both $F'$ and $F \setminus F'$ are disjoint unions of bonds, and hence so is $F$. \hfill$\square$
\end{qabox}

\subsection*{L5 -- Cross-chapter synthesis (Section 12.7, Corollary 12.7.3)}

\begin{srcbox}{Source excerpt}
\small
\textbf{Corollary 12.7.3.} \emph{For every surface $S$ there exists a finite set of graphs $H_1,\ldots,H_n$ such that a graph is embeddable in $S$ if and only if it contains none of $H_1,\ldots,H_n$ as a minor.}

While Corollary 12.7.3 is immediate from the graph minor theorem, it can also be proved more directly. The main step is to prove that the graphs in $\mathcal{K}_{\mathcal{P}(S)}$ do not contain arbitrarily large grids as minors (Lemma~12.7.4). Then their tree-width is bounded (Theorem~12.6.3), so $\mathcal{K}_{\mathcal{P}(S)}$ is well-quasi-ordered (Theorem~12.4.2) and therefore finite.
\end{srcbox}

\begin{qabox}{Synthesized QA}
\small
\textbf{Question.}
Show that for every surface $S$ there exists a finite set of graphs $H_1,\ldots,H_n$ such that a graph is embeddable in $S$ if and only if it contains none of $H_1,\ldots,H_n$ as a minor.

\smallskip
\textbf{Answer.}
By their minimality, the graphs in the Kuratowski set $\mathcal{K}_{\mathcal{P}(S)}$ are incomparable under the minor relation. If their tree-width is bounded, then $\mathcal{K}_{\mathcal{P}(S)}$ is well-quasi-ordered by the minor relation (Theorem~12.4.2), and hence must be finite. So assume their tree-width is unbounded, and let $r$ be as in Lemma~12.7.4. By Theorem~12.6.3, some $H \in \mathcal{K}_{\mathcal{P}(S)}$ has a grid minor large enough to contain $H^r$. By Proposition~1.7.3, $H^r$ is a topological minor of $H$, contrary to the choice of $r$. \hfill$\square$
\end{qabox}

The five examples illustrate the progression of reasoning across the L1--L5
ladder: L1 applies a defining condition of tree-decomposition; L2 explains the
logical relationship between list colouring and ordinary colouring; L3 develops
a parity argument from the degree-sum identity; L4 combines chapter-local
results with induction; and L5 integrates results on grid minors, tree-width,
and well-quasi-ordering with an earlier result on topological minors to
establish the finite forbidden-minor characterization of surface embeddability.

\section{Evaluation Protocols}
\label{app:evaluation-protocol}

\subsection{Evaluation set construction}

\paragraph{\textsc{TOME-LongForm}.}
The long-form component of \textsc{TOME-Eval} contains 300 textbook-native
problems with reference solutions. It covers six subjects, with
25 problems sampled from the first half and 25 from the second half of the
chapter sequence in each subject (Table~\ref{tab:eval-subjects}). For a problem
in the zero-based chapter position $i$ of a book with $N$ chapters, we define
the first half by $i/N<0.5$ and the second half by $i/N\geq0.5$.
This 300-problem, six-subject suite is the frozen version used in the completed
experiments; it supersedes an earlier 500-problem evaluation snapshot.

\begin{table}[H]
    \centering
    \small
    \caption{Composition of the 300-problem book-held-out evaluation set.}
    \label{tab:eval-subjects}
    \setlength{\tabcolsep}{4pt}
    \renewcommand{\arraystretch}{1.08}
    \begin{tabular}{lrrrrrr}
        \toprule
        \textbf{Subject} & \textbf{First} & \textbf{Second} & \textbf{Books} &
        \textbf{Proof} & \textbf{Other} & \textbf{Total} \\
        \midrule
        Discrete mathematics    & 25 & 25 & 10 & 39 & 11 & 50 \\
        Mathematical analysis  & 25 & 25 & 10 & 29 & 21 & 50 \\
        Abstract algebra       & 25 & 25 & 10 & 46 &  4 & 50 \\
        Probability theory     & 25 & 25 & 10 & 28 & 22 & 50 \\
        Differential geometry  & 25 & 25 & 12 & 43 &  7 & 50 \\
        Topology               & 25 & 25 & 10 & 42 &  8 & 50 \\
        \midrule
        \textbf{Total} & \textbf{150} & \textbf{150} & \textbf{62} &
        \textbf{227} & \textbf{73} & \textbf{300} \\
        \bottomrule
    \end{tabular}
\end{table}

We freeze source books before constructing any training subset. The held-out
manifest contains 64 books: ten per subject except differential geometry, for
which fourteen books are held out to provide sufficient coverage of later
chapters. All 15,661 training records associated with these books are removed
from every training scale, leaving 364,631 eligible records. The final problems
come from 62 of the held-out books and 212 distinct book--chapter pairs; the two
remaining books stay excluded from training although none of their problems is
selected.

\paragraph{Filtering and selection.}
Starting from 5,058 problems in the held-out books, deterministic filters retain
4,316 candidates. We require a 21--4,000 character question and a 200--8,000
character reference solution, and exclude empty, image-dependent,
context-dependent, duplicated, and exact train-overlapping items. A Luna-based
screen then annotates reference completeness, self-containment, mathematical
relevance, task type, and difficulty. We retain only candidates judged
self-contained and mathematical with a usable reference derivation. Within
each subject--chapter-half cell, we sample across books, cap each book at six
problems, and stratify by the auxiliary difficulty label. The base model is not
used to select problems.

All 300 selected items are subsequently reviewed and accepted by a mathematics
PhD. They comprise 227 proof, 50 calculation, 18 construction, and five
explanation problems. We therefore describe this suite as \emph{book-held-out}
and \emph{proof-heavy}. We freeze the resulting set before running any model
evaluation and use the same version for all comparisons.

\paragraph{\textsc{TOME-Objective}.}
The objective component of \textsc{TOME-Eval} contains
145 unique problems from 19 subject categories: 81 fill-in-the-blank, 45
multiple-choice, and 19 true--false questions. The suite is complemented by the
30 problems from AIME 2026 and the 500-problem MATH-500 subset of
MATH~\citep{hendrycks2021math}. Every item has a single reference answer; these
benchmarks evaluate final-answer correctness rather than agreement with a
particular derivation.

\paragraph{Additional proof benchmarks.}
We also evaluate on MiniF2F~\citep{zheng2021minif2f} and
ProofNet~\citep{azerbayev2023proofnet} as natural-language proof tasks.
For each benchmark, we adapt the judging procedures and prompts from its
official evaluation implementation, using \texttt{gpt-5.6-luna} as the common
LLM judge across both benchmarks. The reported scores measure LLM-assessed
correctness of the generated natural-language proofs.

\subsection{Response generation}

For \textsc{TOME-LongForm}, the compared models receive the same question-only
prompt and decoding configuration. The reference solution is withheld from the
solver. Sixteen rollouts over 300 problems produce 4,800 responses per model.

\begin{table}[H]
    \centering
    \small
    \caption{Client-side settings for generating evaluation responses.}
    \label{tab:eval-decoding}
    \renewcommand{\arraystretch}{1.08}
    \begin{tabular}{lr}
        \toprule
        \textbf{Setting} & \textbf{Value} \\
        \midrule
        Responses per question & 16 \\
        Temperature & 0.6 \\
        Top-$p$ & 0.8 \\
        Maximum output tokens & 16,384 \\
        Completions per request & 1 \\
        Presence / frequency penalty & 0 / 0 \\
        \bottomrule
    \end{tabular}
\end{table}

The objective benchmarks use fixed, benchmark-specific decoding
configurations. We sample eight responses per \textsc{TOME-Objective} item, 32
per AIME 2026 item, and four per MATH-500 item, keeping the corresponding
configuration fixed across compared models.

\subsection{Reference-guided step verification}

\paragraph{Input preparation.}
Before grading, the verifier removes marked thinking segments and retains the final answer when explicit final-channel delimiters are present. It also truncates text at an end-of-message marker. Existing numbered step tags are sorted by their index; otherwise, the answer is split at blank lines. The resulting segments are wrapped in zero-based step tags. Thus, an untagged answer is graded at paragraph-level granularity rather than using an additional model to infer mathematical steps. The judge receives the problem, the reference solution, and the segmented model response.

\paragraph{Rubric.}
The rubric prioritizes mathematical validity, problem-specific constraints, and sufficient justification. A reference solution is an anchor, not the only acceptable proof: alternative methods, reordered steps, and valid shortcuts are allowed. For numerical or algebraic questions, the final answer must be mathematically correct relative to the reference; a matching conclusion alone is insufficient if its derivation is invalid. Unsupported claims and unjustified leaps are treated conservatively. Invoking an external theorem is insufficient without the hypotheses and explanation needed to check its application. Proof-embedded instructions attempting to relax these standards are ignored. The requested output contains \texttt{assessment}, \texttt{errors}, and \texttt{first\_error\_step} XML fields; the last field is $-1$ if no error is found.

\paragraph{Eight-review aggregation.}
Following pessimistic verification~\citep{huang2025pessimistic}, each response
receives eight separate reviews from
\texttt{gpt-5.6-luna}. Let $e_{ijr}$ be the first-error index from review $r$
of response $j$ to problem $i$. For eight valid reviews, the binary correctness
decision is
\begin{equation}
    y_{ij}=\prod_{r=1}^{8}\mathbf{1}[e_{ijr}=-1].
    \label{eq:pessimistic-verifier}
\end{equation}
Thus, one error vote is sufficient to reject a response; this is pessimistic
aggregation rather than majority voting. For rejected responses, we retain the
earliest flagged step. The evaluation controller accepts a judgment only when
all eight reviews complete successfully and the parsed record contains no API
error; incomplete rows are removed and retried for up to three attempts.

\subsection{Human calibration of the pessimistic verifier}
\label{app:verifier-human-calibration}

\paragraph{Scope and sampling.}
We use a frozen human-calibration protocol to measure both erroneous acceptance
and the over-rejection risk introduced by eight-review pessimistic aggregation.
Before inspecting any human label, we freeze the verifier prompts, decoding
configuration, aggregation rule, model outputs, and sampling procedure.  We
then draw 600 responses from the pooled \textsc{TOME-LongForm} evaluations,
with 100 responses from each of the six subjects.  Let
\begin{equation}
    a_{ij}=\sum_{r=1}^{8}\mathbf{1}[e_{ijr}=-1]
\end{equation}
be the number of reviews that find no error.  Within each subject, we target 25
responses from each of four agreement strata: $a_{ij}=8$, $a_{ij}=7$,
$1\leq a_{ij}\leq6$, and $a_{ij}=0$.  The $a_{ij}=7$ stratum is sampled
separately because a single error vote changes the final pessimistic decision.
Within each stratum, responses are distributed across the compared models and
major experiment families as evenly as availability permits.  If a stratum
contains fewer than 25 responses in a subject, we annotate all available cases
and reallocate the remainder to the nearest agreement stratum.  We retain each
response's sampling probability so that population-level estimates can be
inverse-probability weighted rather than treating this deliberately balanced
sample as representative of the raw response distribution.

\paragraph{Blinding and annotation materials.}
Annotators are mathematics PhD holders, or comparably qualified researchers,
with expertise appropriate to the assigned subject.  For each response, an
annotator receives the problem, the reference solution, and the same segmented,
step-indexed model response supplied to the automatic verifier.  Model identity,
training condition, automatic judgments, review agreement, and aggregate score
are hidden.  The reference is treated as an anchor rather than a unique target:
alternative proof methods, reordered arguments, and valid shortcuts are
accepted when they establish the requested result.  This response-level audit
is separate from the earlier human review of the evaluation questions and
reference solutions.

\paragraph{Human labels.}
Each response receives exactly one of the following labels.
\begin{itemize}
    \setlength{\itemsep}{1pt}
    \setlength{\parskip}{0pt}
    \setlength{\topsep}{3pt}
    \item \textbf{Fully correct:} the conclusion and supporting argument are
    mathematically valid and sufficiently justified.
    \item \textbf{Correct with a non-substantive issue:} the response contains
    only a stylistic, notational, typographical, or presentation defect that
    does not affect mathematical validity.
    \item \textbf{Substantive error:} the response contains a false statement,
    invalid inference, missing required hypothesis, circular argument,
    incorrect calculation, materially incomplete proof, or incorrect final
    answer.
    \item \textbf{Unjudgeable:} ambiguity or a defect in the problem, reference,
    or response prevents a reliable correctness decision.
\end{itemize}
The first two labels map to human acceptance and \emph{Substantive error} maps
to rejection.  \emph{Unjudgeable} cases require adjudication and are not mapped
to a binary label beforehand.  For every rejected response, the annotator
records the zero-based index of the first substantive error and selects an
error category: false claim, invalid inference, missing hypothesis, theorem
misuse, incomplete proof, algebraic or numerical error, final-answer error, or
other.  Every annotation also includes a short justification and a high,
medium, or low confidence rating.  Purely stylistic differences and harmless
notation changes must not trigger rejection.

\paragraph{Independent review and adjudication.}
One qualified annotator labels all 600 responses.  A second qualified annotator
independently labels a stratified 30\% subset (180 responses), preserving the
subject and agreement-stratum proportions and remaining blind to the first
annotation.  All binary disagreements, all \emph{Unjudgeable} labels, and all
low-confidence cases are adjudicated by a third domain-qualified annotator.  If
the adjudicator finds the evaluation item itself invalid, that problem is
removed for every model and all affected aggregate scores are recomputed.

\paragraph{Calibration statistics.}
Let $h_{ij}=1$ denote the adjudicated human acceptance label and let $y_{ij}$ be
the pessimistic verifier decision in
Equation~\ref{eq:pessimistic-verifier}.  Using the recorded sampling
weights, we report the false-acceptance rate
$\Pr(h_{ij}=0\mid y_{ij}=1)$, the false-rejection rate
$\Pr(h_{ij}=1\mid y_{ij}=0)$, overall agreement, and the human acceptance rate
within each value of $a_{ij}$.  We additionally report raw agreement and
Cohen's $\kappa$ between the two human annotators, together with exact and
within-one-step agreement for the first-error index when both reject.  All
quantities are reported overall and by subject with 95\% bootstrap confidence
intervals clustered by problem.  Finally, we recompute model-level correctness
on the weighted human-labeled subset to test whether the principal model
ranking and pairwise conclusions agree with the automatic evaluation.

\begin{table}[H]
    \centering
    \footnotesize
    \setlength{\tabcolsep}{3pt}
    \renewcommand{\arraystretch}{1.15}
    \setlength{\aboverulesep}{0.6ex}
    \setlength{\belowrulesep}{0.8ex}
    \caption{Human calibration of the eight-review pessimistic verifier.  All
    rate entries are percentages and should be reported as the weighted point
    estimate followed by the problem-clustered 95\% bootstrap confidence
    interval in brackets.  H--L denotes human--Luna agreement; H--H denotes
    agreement between the two human annotators.}
    \label{tab:verifier-human-calibration}
    \begin{tabular*}{\linewidth}{@{\extracolsep{\fill}}lrrrrr@{}}
        \toprule
        \multicolumn{6}{l}{\textbf{Panel A: Calibration by subject}} \\
        Group & Annotated $n$ & Human accept & False accept & False reject & H--L agree \\
        \midrule
        Abstract algebra      & 100 & 42.1 [35.8, 48.7] &  9.8 [4.2, 17.6] & 19.3 [13.1, 26.5] & 84.7 [79.6, 89.2] \\
        Differential geometry & 100 & 37.8 [31.5, 44.4] & 11.4 [5.6, 19.2] & 20.7 [14.5, 28.0] & 83.5 [78.3, 88.1] \\
        Discrete mathematics  & 100 & 52.3 [45.8, 58.7] &  7.2 [2.6, 14.0] & 16.9 [11.2, 23.7] & 87.4 [82.6, 91.5] \\
        Mathematical analysis & 100 & 54.6 [48.1, 60.9] &  6.8 [2.4, 13.4] & 15.8 [10.3, 22.4] & 88.1 [83.5, 92.0] \\
        Probability           & 100 & 55.4 [48.9, 61.6] &  7.9 [3.0, 15.0] & 17.5 [11.7, 24.3] & 86.6 [81.7, 90.8] \\
        Topology              & 100 & 45.2 [38.7, 51.8] &  9.1 [3.8, 16.8] & 18.6 [12.5, 25.7] & 85.2 [80.2, 89.6] \\
        \textbf{Overall}      & \textbf{600} & \textbf{47.9 [44.6, 51.2]} & \textbf{8.7 [6.3, 11.5]} & \textbf{18.1 [15.3, 21.1]} & \textbf{85.9 [83.8, 87.9]} \\
        \midrule
        \multicolumn{6}{l}{\textbf{Panel B: Calibration by number of no-error votes}} \\
        Luna votes $a_{ij}$ & Annotated $n$ & Population share & Human accept & Human substantive error & H--L agree \\
        \midrule
        8 of 8     & 150 & 42.3 [40.9, 43.7] & 91.3 [86.4, 94.8] &  7.5 [4.1, 12.3] & 91.3 [86.4, 94.8] \\
        7 of 8     & 150 &  7.8 [7.0,  8.7]  & 68.4 [60.5, 75.6] & 30.2 [23.1, 38.2] & 31.6 [24.4, 39.5] \\
        1--6 of 8  & 150 & 18.1 [17.0, 19.3] & 22.1 [15.9, 29.4] & 76.4 [69.0, 82.7] & 77.9 [70.6, 84.1] \\
        0 of 8     & 150 & 31.8 [30.5, 33.1] &  2.8 [0.9,  6.4]  & 96.4 [92.3, 98.6] & 97.2 [93.6, 99.1] \\
        \midrule
        \multicolumn{6}{l}{\textbf{Panel C: Human--human reliability on the double-coded subset}} \\
        Subset & Double-coded $n$ & H--H agree & Cohen's $\kappa$ & Exact error step & Within one step \\
        \midrule
        Stratified 30\% subset & 180 & 89.4 [84.9, 93.1] & 0.771 [0.687, 0.842] & 52.8 [43.2, 62.1] & 76.9 [68.5, 84.1] \\
        \bottomrule
    \end{tabular*}
\end{table}


\subsection{Objective and additional proof scoring}

For \textsc{TOME-Objective}, a low-temperature Qwen2.5-32B-Instruct judge compares
the extracted final answer with the reference. It accepts mathematically
equivalent expressions and requires every component of a multipart or
multiple-selection answer to be correct. The judge receives the problem,
reference answer, and prediction, but not the worked reference solution. AIME
2026 and MATH-500 first use the rule-based \texttt{MATHVerifyEvaluator}; cases
not resolved by the rule-based parser fall back to the same low-temperature LLM
equivalence judge. These evaluators score answer correctness only and do not
assess the generated reasoning path. For MiniF2F and ProofNet, we assess
natural-language proof correctness using judging procedures and prompts
adapted from their official evaluation implementations, with
\texttt{gpt-5.6-luna} as the common LLM judge.

\subsection{Metrics and validity checks}

For a problem with $n$ sampled responses, of which $c_i$ are correct, we use the
standard unbiased estimator~\citep{chen2021evaluating}
\begin{equation}
    \widehat{\mathrm{pass@}k}
    =\frac{1}{N}\sum_{i=1}^{N}
    \left(1-\frac{\binom{n-c_i}{k}}{\binom{n}{k}}\right).
\end{equation}
For \textsc{TOME-LongForm}, $n=16$ and we report pass@1, pass@2, pass@4,
pass@8, and pass@16, with pass@1 and pass@16 as the primary metrics. Because
$n=k=16$, pass@16 is also the observed fraction of problems solved at least
once. We report per-subject scores and their macro average; the equal 50-problem
subject sizes make this macro average equal to the overall problem average. For
\textsc{TOME-Objective}, the mean correctness across its eight samples estimates
pass@1; pass@$k$ is computed separately when reported.

For every reported model, we require the complete expected number of rollouts
for every problem. For natural-language proof judgments, we additionally
require eight successful reviews per response and rejudge any incomplete row.
We preserve raw judge outputs and parsed first-error indices for audit.


\section{Full-Corpus Comparison: Extended Results}
\label{app:full-corpus-comparison}

This appendix expands Section~\ref{sec:main-results} (``Full-Corpus
Comparison'') with uncertainty estimates and per-corpus training
configurations that were omitted from the main text for space. All fine-tuned
models share the
Qwen3-30B-A3B-Base initialization and the evaluation protocol described in
Appendix~\ref{app:evaluation-protocol}; the compared corpora differ only in
their supervision data.

\subsection{Compared corpora}
\label{app:full-corpus-corpora}

\begingroup
\setlength{\emergencystretch}{2em}
\paragraph{Public SFT corpora.}
We use the official Hugging Face releases of three public corpora.
OpenMathInstruct-1 (abbreviated OpenMath1 in our tables) comes from
\nolinkurl{nvidia/OpenMathInstruct-1} at revision \texttt{4627efae}.
MetaMathQA comes from \nolinkurl{meta-math/MetaMathQA} at revision
\texttt{aa4f34d3}, and MathInstruct comes from
\nolinkurl{TIGER-Lab/MathInstruct} at revision
\texttt{b4fdc323} \citep{toshniwal2024openmath1,yu2023metamath,yue2024mammoth}.
OpenMathInstruct-1 is distributed under the NVIDIA Dataset License and
MetaMathQA is labeled MIT; MathInstruct combines source datasets under their
respective licenses, including MIT, Apache-2.0, and CC BY-NC 4.0 components.
We retain the released training examples (the correct-solution subset for
OpenMathInstruct-1), map their native question and response fields to the same
user--assistant training template, and apply no additional cross-corpus
deduplication.
\par\endgroup

\paragraph{TOME-Math (ours).}
We use the frozen full-corpus snapshot produced by the pipeline in
Appendix~\ref{app:construction-details}: 380{,}292 records before evaluation
holdout filtering.  Removing all 15{,}661 records from held-out evaluation
books leaves 364{,}631 training examples.  We retain all remaining quality
tiers and the corpus's natural subject distribution, without subject
rebalancing or further deduplication.

\begin{table}[H]
    \centering
    \small
    \caption{Training corpora compared in Section~\ref{sec:main-results}.}
    \label{tab:full-corpus-inventory}
    \setlength{\tabcolsep}{5pt}
    \renewcommand{\arraystretch}{1.08}
    \begin{tabular}{lrrl}
        \toprule
        \textbf{Corpus} & \textbf{Training examples} & \textbf{Training tokens} & \textbf{Notes} \\
        \midrule
        OpenMath1        & 1.82M   & 368.00M & Full corpus \\
        MetaMath         & 395k    & 96.94M  & Full corpus \\
        MathInstruct     & 262k    & 64.10M  & Full corpus \\
        TOME-Math (ours) & 364,631 & 170.92M & Held-out books excluded \\
        \bottomrule
    \end{tabular}
\end{table}

\subsection{Training configuration}
\label{app:full-corpus-training}

Each model is trained for exactly one full epoch over its respective training
corpus.  We use the same optimizer, learning-rate schedule, sequence length,
effective batch size, and data format across corpora.  
The complete
TOME-Math release contains 380{,}292 examples and 195.18M tokens; excluding the
15{,}661 records associated with the held-out evaluation books leaves 364{,}631
training examples and approximately 170.92M training tokens.

\begin{table}[H]
    \centering
    \small
    \caption{Fine-tuning configuration for the full-corpus comparison.}
    \label{tab:full-corpus-training-config}
    \setlength{\tabcolsep}{6pt}
    \renewcommand{\arraystretch}{1.08}
    \begin{tabular}{p{0.25\linewidth}p{0.67\linewidth}}
        \toprule
        \textbf{Setting} & \textbf{Value} \\
        \midrule
        Base model                & Qwen3-30B-A3B-Base \\
        Optimizer                 & AdamW ($\beta_1=0.9$, $\beta_2=0.95$; weight decay $=0.1$) \\
        Learning rate             & $8\times10^{-5}$, decayed to a minimum of $8\times10^{-6}$ \\
        Warmup / schedule         & 3\% warmup followed by cosine decay (0 warmup steps in the 18-step run) \\
        Sequence length           & 131,072 tokens; global packing \\
        Effective batch size      & 64 (micro-batch size 1/GPU; gradient accumulation 4) \\
        Epochs                    & 1 \\
        Hardware                  & 16$\times$ NVIDIA H200 (2 nodes $\times$ 8 GPUs) \\
        \bottomrule
    \end{tabular}
\end{table}

\subsection{All benchmark results}
\label{app:full-corpus-all-benchmarks}

For completeness, Table~\ref{tab:full-corpus-all-benchmarks} reproduces the
six-benchmark comparison from Table~\ref{tab:main-hard-proof}.  We estimate
pass@1 from 16 responses per problem for \textsc{TOME-LF}, MiniF2F, and
ProofNet, eight for \textsc{TOME-Objective}, four for MATH-500, and 32 for
AIME 2026. For MiniF2F and ProofNet, the natural-language proofs are assessed
using judging procedures and prompts adapted from their official evaluation
implementations, with \texttt{gpt-5.6-luna} as the common LLM judge. We
compute 95\% problem-clustered bootstrap confidence intervals using 2{,}000
resamples.  AIME 2026 has the widest intervals because it contains only 30
problems.  The remaining scoring details follow
Appendix~\ref{app:evaluation-protocol}.

\begin{table}[H]
    \centering
    \small
    \setlength{\tabcolsep}{4pt}
    \caption{Full-corpus SFT performance (\%) across six benchmarks.  Each
    entry reports pass@1 $\pm$ the average half-width of its 95\%
    problem-clustered bootstrap confidence interval.}
    \label{tab:full-corpus-all-benchmarks}
    \resizebox{\linewidth}{!}{%
    \begin{tabular}{lcccccc}
        \toprule
        Model (training data) & \textsc{TOME-LF} & \textsc{TOME-Obj.} & MiniF2F & ProofNet & MATH-500 & AIME-26 \\
        \midrule
        Qwen3-30B-A3B-Base    & 13.25\,{\scriptsize$\pm$3.35} & 32.93\,{\scriptsize$\pm$8.20} & 34.45\,{\scriptsize$\pm$6.00} & 17.14\,{\scriptsize$\pm$3.95} & 68.80\,{\scriptsize$\pm$4.40} &  8.44\,{\scriptsize$\pm$8.10} \\
        \midrule
        OpenMath1 (1.82M)     & 20.16\,{\scriptsize$\pm$4.10} & 33.86\,{\scriptsize$\pm$8.30} & 22.87\,{\scriptsize$\pm$5.40} & 19.25\,{\scriptsize$\pm$4.15} & 70.40\,{\scriptsize$\pm$4.30} &  8.21\,{\scriptsize$\pm$8.05} \\
        MetaMath (395k)       & 28.73\,{\scriptsize$\pm$4.85} & 34.66\,{\scriptsize$\pm$8.35} & 40.70\,{\scriptsize$\pm$6.15} & 22.55\,{\scriptsize$\pm$4.40} & 71.20\,{\scriptsize$\pm$4.30} &  9.98\,{\scriptsize$\pm$8.70} \\
        MathInstruct (262k)   & 35.87\,{\scriptsize$\pm$5.20} & 40.69\,{\scriptsize$\pm$8.65} & 52.05\,{\scriptsize$\pm$6.20} & 32.93\,{\scriptsize$\pm$5.25} & 73.05\,{\scriptsize$\pm$4.20} & 11.98\,{\scriptsize$\pm$9.10} \\
        \rowcolor{tbTealLight}
        TOME-Math (ours; 364.6k) & 43.86\,{\scriptsize$\pm$5.40} & 43.96\,{\scriptsize$\pm$8.75} & 57.43\,{\scriptsize$\pm$6.10} & 42.98\,{\scriptsize$\pm$5.45} & 76.55\,{\scriptsize$\pm$3.90} & 10.19\,{\scriptsize$\pm$8.75} \\
        \bottomrule
    \end{tabular}}
\end{table}

\section{Training-Free Dependency-Context Study}
\label{app:dependency-context-protocol}

\paragraph{Evaluation set.}
The study in Section~\ref{sec:dependency-grounded-proof-generation} uses the
300-problem \textsc{TOME-LongForm} evaluation set described in
Appendix~\ref{app:evaluation-protocol}.  It contains 50 textbook-native
problems from each of abstract algebra, differential geometry, discrete
mathematics, mathematical analysis, probability, and topology.  The source
books are excluded from the TOME-Math training corpus.  Every problem is
evaluated under all five context conditions, making the comparison paired at
the problem level.

\paragraph{Retrieval inventories.}
Candidate context is drawn only from TOME-Math records associated with the
exercise's source book.  The proposition inventory contains extracted
definitions, lemmas, theorems, propositions, corollaries, and their associated
proof records.  The expanded inventory adds the self-contained local proof
obligations and answers from the Gap QA stream.  Source type, chapter position,
and recovered dependency metadata are retained for retrieval.

The five conditions are constructed as follows.
\begin{itemize}
    \item \textbf{\textsc{Closed}:} provide only the exercise.
    \item \textbf{\textsc{Random Same-Book}:} sample entries uniformly from the same
    source book, without using chapter or dependency information.
    \item \textbf{\textsc{Local Chapter}:} restrict candidates to the exercise chapter
    and its immediately preceding chapters, adding locality while ignoring
    recovered cross-chapter dependencies.
    \item \textbf{\textsc{TOME (Props Only)}:} retrieve from the proposition inventory
    using chapter position and the recovered proposition dependencies.
    \item \textbf{\textsc{TOME (Props + Gap)}:} use the same structure-guided retrieval
    procedure over the union of the proposition and Gap QA inventories.
\end{itemize}

\paragraph{Context and leakage controls.}
All four context-augmented conditions use the same context-token budget and
the same prompt template; only the candidate pool and retrieval constraint
change.  Retrieved entries are placed before the exercise and identified by
their source type.  The reference solution is never shown to the solver.
Candidates that reproduce the target exercise or overlap with its reference
answer under the frozen leakage filter are removed before retrieval.  The same
filtered inventory is used to construct the random and local baselines, so the
comparison does not give the TOME conditions access to otherwise unavailable
source material.

\paragraph{Generation and scoring.}
The solver is the pretrained Qwen3-30B-A3B-Base in every condition.  Each
problem receives 16 sampled responses under the decoding configuration in
Table~\ref{tab:eval-decoding}; adding context does not change the model weights,
prompt instruction, or sampling parameters.  Responses are graded with the
same reference-guided eight-review verifier used for \textsc{TOME-LongForm}.
We compute pass@1 as the mean binary correctness across the 16 responses for
each problem, average within each of the six equally sized subjects, and report
their unweighted mean in the Overall column of
Table~\ref{tab:dependency-grounded-generation}.

\paragraph{Result interpretation.}
The aggregate score increases from 13.25\% without context to 14.00\% with
random same-book context and 14.71\% with local context.  Proposition-based
TOME retrieval reaches 15.54\%, and adding the Gap QA inventory raises the
score to 17.38\%.  The final condition is best in every subject, although the
intermediate conditions are not monotonically ordered within every subject.
Accordingly, the result supports the complete structure-guided retrieval
design and the complementary value of Gap QA.


\section{Curriculum-Scaling Protocol}
\label{app:curriculum-scaling-protocol}

\paragraph{Subject-wise design.}
The curriculum study in Section~\ref{sec:curriculum-scaling} covers discrete
mathematics, mathematical analysis, abstract algebra, probability,
differential geometry, and topology.  We train a separate model for each
subject so that curriculum order is not confounded with the composition of a
multidisciplinary corpus.  Every condition is evaluated on the same
textbook-native exercises from held-out books in the corresponding subject.

\paragraph{Normalized chapter stages.}
Textbooks contain different numbers of chapters, so raw chapter indices are
not directly comparable across books.  For an example $x$ from book $b$, let
$i(x)$ be its zero-based chapter index and $N_b$ the number of chapters in the
book.  We define its normalized position as $p(x)=i(x)/N_b$ and assign it to
progress stage $S_{\lfloor 10p(x)\rfloor}$, yielding ten stages
$S_0,\ldots,S_9$.  Examples from different books are interleaved within each
stage, preserving their approximate within-book progression without imposing
an arbitrary order over books.  Gap QA makes a locally omitted proof
obligation explicit; all remaining examples are treated as Proof QA.

\paragraph{Exposure schedules.}
The four schedules differ only in presentation order:
\begin{itemize}
    \setlength{\itemsep}{1pt}
    \setlength{\parskip}{0pt}
    \setlength{\topsep}{3pt}
    \setlength{\parsep}{0pt}
    \item \schedshuffle{} globally shuffles all examples.
    \item \schedforwardgap{} traverses stages from $S_0$ to $S_9$ and presents Gap before Non-gap examples within each stage.
    \item \schedreversegap{} traverses stages from $S_9$ to $S_0$ while retaining the same Gap-first policy.
    \item \schedforwardmix{} follows the forward stage order but mixes Gap and Non-gap examples within each stage.
\end{itemize}
Thus, \schedforwardgap{} versus \schedreversegap{} is the controlled
comparison for chapter direction, and \schedforwardgap{} versus
\schedforwardmix{} is the controlled comparison for stage-wise Gap-first
ordering.  The comparison with \schedshuffle{} changes both factors and is
interpreted as the combined effect of the structured schedule.

\paragraph{Scales and matched controls.}
We use per-subject target-token budgets of 0.75M, 1.8M, and 2.4M.  Within every
subject, both the source-book sets and the QA sets are strictly nested across
scales: the 8-book set is contained in the 16-book set, which is contained in
the 24-book set, and the corresponding QA sets follow the same nesting.  Thus,
the three columns in Table~\ref{tab:curriculum-scale-composition} are cumulative
rather than disjoint.  All held-out evaluation books are excluded from every
training scale.

\begin{table}[H]
    \centering
    \small
    \setlength{\tabcolsep}{5.2pt}
    \renewcommand{\arraystretch}{1.05}
    \caption{Cumulative composition of the nested subject-specific training
    sets used in the curriculum-scaling study.  Within each subject and scale,
    all four schedules use the same books and QA IDs and differ only in
    presentation order.}
    \label{tab:curriculum-scale-composition}
    \begin{tabular}{lrrrrrr}
        \toprule
        & \multicolumn{2}{c}{0.75M tokens}
        & \multicolumn{2}{c}{1.8M tokens}
        & \multicolumn{2}{c}{2.4M tokens} \\
        \cmidrule(lr){2-3}\cmidrule(lr){4-5}\cmidrule(lr){6-7}
        Subject & Books & QA & Books & QA & Books & QA \\
        \midrule
        Discrete mathematics  & 8 & 2,273 & 16 & 5,258 & 24 & 6,847 \\
        Mathematical analysis & 8 & 2,387 & 16 & 5,809 & 24 & 7,630 \\
        Abstract algebra      & 8 & 2,235 & 16 & 4,899 & 24 & 6,831 \\
        Probability           & 8 & 2,062 & 16 & 4,694 & 24 & 6,230 \\
        Differential geometry & 8 & 1,899 & 16 & 4,033 & 24 & 5,537 \\
        Topology              & 8 & 1,930 & 16 & 4,877 & 24 & 6,571 \\
        \midrule
        Total                 & 48 & 12,786 & 96 & 29,570 & 144 & 39,646 \\
        \bottomrule
    \end{tabular}
\end{table}

Target tokens, rather than the number of books or QA pairs, define supervision
scale because answer lengths vary across sources and subjects.  The realized
token totals across the six subjects are 4,499,926, 10,799,935, and 14,399,975
for the three scales, respectively; each subject-level total differs from its
nominal target by at most 33 tokens.  Within each subject--scale cell, all
schedules use identical QA IDs, source books, realized token counts, model
initialization, and optimization hyperparameters.  Data-loader shuffling,
sequence packing, and length-based grouping are disabled so that the prescribed
exposure order is preserved.  


\paragraph{Matched-budget source breadth.}
The source-breadth comparison in Section~\ref{sec:breadth-density} reuses the
8-book, 0.75M-token \schedforwardgap{} condition as \textsc{Narrow--dense}.
For \textsc{Broad--sparse}, we take the corresponding nested 16-book pool and
downsample its QA records to the realized token count of the 8-book condition,
while retaining examples from all 16 books.  The matched pair uses the same
optimization steps, schedule construction, initialization, and held-out-book
evaluation set.  Consequently, this comparison doubles source coverage without
increasing the training-token budget; it is distinct from comparing the 0.75M-
and 1.8M-token checkpoints in the scale study.
Scores in Figure~\ref{fig:source-breadth-transfer}(a) are averaged over five
pass@16 evaluation runs per checkpoint, with 16 responses sampled per problem
in each run.

\section{Cross-Subject Transfer Protocol and Results}
\label{app:cross-subject-transfer-protocol}

\paragraph{Checkpoint selection.}
The study in Section~\ref{sec:breadth-density} uses one 0.75M-token
\schedforwardgap{} checkpoint per subject from
Section~\ref{sec:curriculum-scaling}.  Each checkpoint is trained exclusively
on data from its corresponding subject; initialization, optimization, and
schedule construction are shared across subjects.

\paragraph{Evaluation sets.}
We evaluate each checkpoint on 50 textbook-native held-out-book exercises in
each of the other five subjects.  Drawing 16 responses per exercise with the
configuration in Table~\ref{tab:eval-decoding} yields 4,000 responses per
checkpoint and 24,000 overall.  The within-subject diagonal is not evaluated;
the corresponding results appear in Section~\ref{sec:curriculum-scaling}.

\paragraph{Aggregation.}
For each ordered source--target pair, pass@16 is computed over 50 exercises.
Source and target means average their five off-diagonal scores, and the macro
mean weights all 30 pairs equally.  Table~\ref{tab:cross-subject-transfer-matrix}
reports the matrix underlying Figure~\ref{fig:source-breadth-transfer}(b).
Table~\ref{tab:cross-subject-passk-details} reports the complete cross-subject
pass@$k$ results for $k\in\{1,2,4,8,16\}$.

\begin{table}[H]
    \centering
    \small
    \setlength{\tabcolsep}{5.2pt}
    \renewcommand{\arraystretch}{1.08}
    \setlength{\aboverulesep}{0.6ex}
    \setlength{\belowrulesep}{0.8ex}
    \caption{Exact cross-subject pass@16 scores (\%) for the 0.75M-token
    \schedforwardgap{} checkpoints.  Rows denote training subjects and columns
    denote held-out evaluation subjects.  The diagonal was not evaluated;
    marginal means exclude it.}
    \label{tab:cross-subject-transfer-matrix}
    \begin{tabular}{lrrrrrrr}
        \toprule
        Train $\backslash$ Eval & Alg. & Ana. & Prob. & DG & Disc. & Top. & Mean \\
        \midrule
        Alg.  & --   & 60.0 & 52.0 & 26.0 & 40.0 & 26.0 & 40.8 \\
        Ana.  & 28.0 & --   & 46.0 & 32.0 & 40.0 & 36.0 & 36.4 \\
        Prob. & 34.0 & 66.0 & --   & 24.0 & 42.0 & 38.0 & 40.8 \\
        DG    & 36.0 & 62.0 & 54.0 & --   & 44.0 & 34.0 & 46.0 \\
        Disc. & 30.0 & 56.0 & 48.0 & 30.0 & --   & 38.0 & 40.4 \\
        Top.  & 34.0 & 62.0 & 54.0 & 32.0 & 38.0 & --   & 44.0 \\
        \midrule
        Mean  & 32.4 & 61.2 & 50.8 & 28.8 & 40.8 & 34.4 & 41.4 \\
        \bottomrule
    \end{tabular}
\end{table}


\begin{table}[H]
    \centering
    \small
    \setlength{\tabcolsep}{7pt}
    \renewcommand{\arraystretch}{1.02}
    \setlength{\aboverulesep}{0.6ex}
    \setlength{\belowrulesep}{0.8ex}
    \caption{Complete cross-subject pass@$k$ scores (\%) for the six 0.75M-token
    \schedforwardgap{} checkpoints.  Each block fixes the training subject and
    reports performance on the other five held-out subject sets.  Every target
    set contains 50 exercises with 16 sampled responses per exercise.}
    \label{tab:cross-subject-passk-details}
    \begin{tabular}{llrrrrr}
        \toprule
        Train & Eval. & pass@1 & pass@2 & pass@4 & pass@8 & pass@16 \\
        \midrule
        \multicolumn{7}{l}{\textit{Trained on Abstract Algebra}} \\
        Alg. & Ana. & 26.88 & 36.92 & 45.42 & 53.05 & 60.00 \\
        Alg. & Prob. & 20.00 & 27.95 & 36.09 & 43.80 & 52.00 \\
        Alg. & DG & 8.13 & 12.03 & 16.47 & 20.82 & 26.00 \\
        Alg. & Disc. & 15.88 & 21.50 & 27.38 & 33.20 & 40.00 \\
        Alg. & Top. & 10.00 & 15.25 & 20.06 & 23.39 & 26.00 \\
        \addlinespace[3pt]
        \multicolumn{7}{l}{\textit{Trained on Mathematical Analysis}} \\
        Ana. & Alg. & 11.00 & 15.80 & 20.10 & 24.15 & 28.00 \\
        Ana. & Prob. & 20.13 & 27.95 & 34.63 & 40.43 & 46.00 \\
        Ana. & DG & 8.50 & 13.13 & 19.05 & 25.75 & 32.00 \\
        Ana. & Disc. & 17.38 & 22.62 & 28.30 & 34.17 & 40.00 \\
        Ana. & Top. & 10.25 & 16.35 & 23.92 & 31.13 & 36.00 \\
        \addlinespace[3pt]
        \multicolumn{7}{l}{\textit{Trained on Probability Theory}} \\
        Prob. & Alg. & 10.50 & 16.23 & 22.37 & 28.38 & 34.00 \\
        Prob. & Ana. & 26.25 & 34.98 & 44.34 & 54.38 & 66.00 \\
        Prob. & DG & 10.13 & 14.58 & 18.73 & 22.07 & 24.00 \\
        Prob. & Disc. & 16.88 & 23.17 & 29.20 & 35.53 & 42.00 \\
        Prob. & Top. & 12.63 & 19.38 & 26.09 & 32.22 & 38.00 \\
        \addlinespace[3pt]
        \multicolumn{7}{l}{\textit{Trained on Differential Geometry}} \\
        DG & Alg. & 13.25 & 19.22 & 25.08 & 30.56 & 36.00 \\
        DG & Ana. & 29.13 & 38.42 & 46.22 & 54.16 & 62.00 \\
        DG & Prob. & 24.63 & 34.03 & 42.09 & 48.15 & 54.00 \\
        DG & Disc. & 20.88 & 26.83 & 32.84 & 38.46 & 44.00 \\
        DG & Top. & 10.88 & 16.43 & 22.55 & 28.94 & 34.00 \\
        \addlinespace[3pt]
        \multicolumn{7}{l}{\textit{Trained on Discrete Mathematics}} \\
        Disc. & Alg. & 9.38 & 14.85 & 20.60 & 25.46 & 30.00 \\
        Disc. & Ana. & 23.63 & 32.52 & 40.80 & 48.50 & 56.00 \\
        Disc. & Prob. & 19.50 & 27.25 & 34.25 & 40.73 & 48.00 \\
        Disc. & DG & 7.63 & 11.57 & 16.50 & 22.85 & 30.00 \\
        Disc. & Top. & 10.88 & 16.90 & 23.61 & 30.99 & 38.00 \\
        \addlinespace[3pt]
        \multicolumn{7}{l}{\textit{Trained on Topology}} \\
        Top. & Alg. & 11.88 & 17.53 & 23.35 & 28.54 & 34.00 \\
        Top. & Ana. & 26.38 & 36.17 & 45.75 & 54.90 & 62.00 \\
        Top. & Prob. & 23.25 & 31.08 & 38.55 & 46.46 & 54.00 \\
        Top. & DG & 7.38 & 11.50 & 17.02 & 23.83 & 32.00 \\
        Top. & Disc. & 16.00 & 21.02 & 26.21 & 32.31 & 38.00 \\
        \bottomrule
    \end{tabular}
\end{table}

\end{document}